\documentclass[10pt]{article}
\usepackage[margin=1in]{geometry}

\usepackage{amsmath} 
\usepackage{amssymb,amsfonts,bm,mathrsfs}
\usepackage{nicematrix}
\usepackage{aligned-overset}
\usepackage{subcaption}
\usepackage[colorlinks = true,
            linkcolor = blue,
            citecolor = gray,
            urlcolor = black,
            ]{hyperref}
\hypersetup{
    linktoc=all, %
}
\usepackage{tikz,pgfplots,float}
\pgfplotsset{compat=newest}
\usetikzlibrary {datavisualization.formats.functions}
\pgfplotsset{compat=1.17}
\usepgfplotslibrary{groupplots}
\usetikzlibrary{matrix}
\usepackage{adjustbox}

\usepackage{amsthm}
\usepackage[nameinlink,capitalize]{cleveref}
\theoremstyle{plain}
\newtheorem{theorem}{Theorem}
\newtheorem{proposition}[theorem]{Proposition}
\newtheorem{lemma}[theorem]{Lemma}
\newtheorem{corollary}[theorem]{Corollary}
\newtheorem*{conjecture*}{Conjecture}

\newtheorem{definition}[theorem]{Definition}

\newtheorem{property}[theorem]{Property}
\renewenvironment{proof}{\noindent{\bfseries Proof.}}{\hfill $\blacksquare$\\}
\allowdisplaybreaks

\usepackage{algorithm}
\usepackage[noend]{algpseudocode}

\DeclarePairedDelimiter\paren{\lparen}{\rparen}
\DeclarePairedDelimiter\abs{\lvert}{\rvert}
\DeclarePairedDelimiter\norm{\lVert}{\rVert}

\DeclarePairedDelimiter\braces{\lbrace}{\rbrace}
\DeclarePairedDelimiter\bracks{\lbrack}{\rbrack}
\let\angle\Angle
\DeclarePairedDelimiter\angle{\langle}{\rangle}

\newcommand{\absm}[1]{\bigl\lvert#1\bigr\rvert}

\newcommand{\parenm}[1]{\biggl\lparen#1\biggr\rparen}
\newcommand{\bracesm}[1]{\biggl\lbrace#1\biggr\rbrace}
\newcommand{\bracksm}[1]{\biggl\lbrack#1\biggr\rbrack}

\def\eps{{\epsilon}}
\newcommand{\defn}{\mathrm{def}}
\newcommand{\as}{\mathrm{a.s.}}
\DeclareMathOperator*{\argmax}{arg\,max}
\DeclareMathOperator*{\argmin}{arg\,min}

\DeclareMathOperator*{\E}{\mathbf{E}}
\DeclareMathOperator*{\Prob}{\mathbf{Pr}}
\DeclareMathOperator*{\Var}{\mathbf{Var}}

\newcommand{\relu}{\textsf{ReLU}}
\newcommand{\sign}{\text{sign}}

\newcommand{\SA}{\mathsf{S}}
\newcommand{\e}{\mathrm{e}}

\def\ermP{{\textnormal{P}}}
\def\ermQ{{\textnormal{Q}}}

\def\ermS{{\textnormal{S}}}

\def\ermX{{\textnormal{X}}}

\def\vzero{{\bm{0}}}
\def\vone{{\bm{1}}}

\def\vxi{{\bm{\xi}}}
\def\vzeta{{\bm{\zeta}}}
\def\va{{\mathbf{a}}}
\def\vb{{\mathbf{b}}}

\def\ve{{\mathbf{e}}}

\def\vg{{\mathbf{g}}}

\def\vu{{\mathbf{u}}}
\def\vv{{\mathbf{v}}}
\def\vw{{\mathbf{w}}}
\def\vx{{\mathbf{x}}}
\def\vy{{\mathbf{y}}}

\def\mA{{A}}
\def\mB{{B}}
\def\mC{{C}}

\def\mE{{E}}

\def\mG{{G}}

\def\mI{{I}}

\def\mS{{S}}
\def\mT{{T}}
\def\mU{{U}}

\def\mW{{W}}
\def\mX{{X}}
\def\mY{{Y}}

\def\mSigma{{\Sigma}}

\def\mGamma{{\Gamma}}

\DeclareMathAlphabet{\mathsfit}{\encodingdefault}{\sfdefault}{m}{sl}
\SetMathAlphabet{\mathsfit}{bold}{\encodingdefault}{\sfdefault}{bx}{n}

\def\gB{{\mathcal{B}}}

\def\gD{{\mathcal{D}}}

\def\gL{{\mathcal{L}}}

\def\gN{{\mathcal{N}}}

\def\gP{{\mathcal{P}}}

\def\gR{{\mathcal{R}}}
\def\gS{{\mathcal{S}}}

\def\gW{{\mathcal{W}}}

\def\sN{{\mathbb{N}}}

\def\sR{{\mathbb{R}}}
\def\sS{{\mathbb{S}}}

\newcommand{\vect}{\mathrm{vec}}

\newcommand{\Proj}{\mathrm{Proj}}

\newcommand{\normsm}[1]{\bigl\lVert#1\bigr\rVert}

\newcommand{\norms}[1]{\lVert#1\rVert}

\DeclareMathOperator{\F}{F}
\DeclareMathOperator{\T}{\top}

\DeclareMathOperator{\Tr}{tr}

\DeclareMathOperator{\poly}{\mathsf{poly}}
\DeclareMathOperator{\polylog}{\mathsf{polylog}}

\newcommand{\lip}{\mathrm{lip}}
\newcommand{\TP}{\mathrm{TP}}
\newcommand{\FP}{\mathrm{FP}}
\newcommand{\TN}{\mathrm{TN}}
\newcommand{\FN}{\mathrm{FN}}
\newcommand{\HT}{\mathrm{HT}}
\newcommand{\sigmoid}{\sigma}

\usepackage{natbib}

\usepackage[colorinlistoftodos,prependcaption,textsize=small]{todonotes}

\begin{document}
\sloppy
\begin{titlepage}
	\def\thepage{}
	\thispagestyle{empty}
	\title{Iterative Thresholding for Non-Linear Learning in the\\ Strong $\eps$-Contamination Model}
	\author{Arvind Rathnashyam\\RPI Math, \href{mailto:rathna@rpi.edu}{\textcolor{black}{\texttt{rathna@rpi.edu}}} \and Alex Gittens\\RPI CS, \href{mailto:gittea@rpi.edu}{\textcolor{black}{\texttt{gittea@rpi.edu}}}}
	\date{}
 	\maketitle
 	\begin{abstract}
	We derive approximation bounds for learning single neuron models using thresholded gradient descent when both the labels and the covariates are possibly corrupted adversarially. We assume $(\vx, y) \sim \gD$ satisfy
    $$
        y = \sigmoid\paren{\vw^{*} \cdot \vx} + \xi,
    $$
    where $\sigmoid$ is a nonlinear activation function, the noise $\xi$ is sampled from $\gN(0,\nu^2)$, and the covariate vector $\vx$ is sampled from a sub-Gaussian distribution. We study sigmoidal, leaky-$\relu$, and $\relu$ activation functions and derive a $O(\nu\sqrt{\epsilon\log(1/\epsilon)})$ approximation bound in $\ell_{2}$-norm, with sample complexity $O(d/\epsilon)$ and failure probability $e^{-\Omega(d)}$. 
    
    We also study the linear regression problem, where $\sigmoid(x) = x$. We derive a $O(\nu\epsilon\log(1/\epsilon))$ approximation bound, improving upon the previous $O(\nu)$ approximation bounds for the gradient-descent based iterative thresholding algorithms of Bhatia et al. (NeurIPS 2015) and Shen and Sanghavi (ICML 2019). Our algorithm has a $O(\polylog(N,d)\log(R/\epsilon))$ runtime complexity when $\norm{\vw^{*}}_{2} \leq R$, improving upon the $O(\polylog(N,d)/\epsilon^2)$ runtime complexity of Awasthi et al. (NeurIPS 2022). 
\end{abstract}
\end{titlepage}

\section{Introduction}
The learning of the parameters of Generalized Linear Models (GLMs) \citep{awasthi:2022,diakonikolas:2019,fischler:1981,li:2021,osama:2020} and linear regression models \citep{bhatia:2017,mukhoty:2019} under the Huber $\eps$-contamination model is well-studied. Efficient algorithms for robust statistical estimation have been studied extensively for problems such as high-dimensional mean estimation \citep{cheng:2020,prasad:2019} and Robust Covariance Estimation \citep{cheng:2019,fan:2018}; see \citet{diakonikolas:2023} for an overview. Along these lines, interest has developed in the development of robust gradient-descent based approaches to machine learning problems \citep{diakonikolas:2019,prasad:2018}. This work advances this line of research by providing gradient-descent based approaches for learning single neuron models under the strong contamination model.

\begin{definition}[Strong $\epsilon$-Contamination Model]
    Given a corruption parameter $0 \leq \eps < 0.5$, and data samples $(\vx_{i},y_{i})$ for $ i \in [N]$, an adversary is allowed to inspect all samples and modify $\epsilon N$ samples arbitrarily. We define the corrupted dataset $\ermX = \ermP \cup \ermQ$ where $\ermP$ contains the remaining target samples and $\ermQ$ contains the samples modified by the adversary. The algorithm is then given the $\eps$-corrupted dataset $\ermX$ as training data. 
\end{definition}

Current approaches for robust learning across various machine learning tasks often use gradient descent over a robust objective (see e.g. Tilted Empirical Risk Minimization (TERM) \citep{li:2021}). These robust objectives tend to not be convex and therefore are difficult to obtain strong approximation bounds for general classes of models. Another popular approach is filtering, where at each iteration of training, points deemed to be as outliers are removed from training (see e.g. SEVER \citep{diakonikolas:2019}). However, filtering algorithms such as SEVER require careful parameter selection to be useful in practice. 

Iterative Thresholding is a popular framework for robust statistical estimation that was introduced in the 19th century by Legendre \citep{legendre1806nouvelles}. Part of its appeal lies in its simplicity, as at each iteration of training we simply ignore points with error above a certain threshold. Despite the venerablity of this approach, it was not until recently that some iterative thresholding algorithms were proven to deliver robust parameter estimates in polynomial time. One of the first theoretical results in the literature proving the effectiveness of iterative thresholding considers its use in estimating the parameters of regression models under additive adversarial corruptions.

\begin{theorem}[Theorem 5 in \cite{bhatia2015robust}]\label{thm:bhatia-thm-5}
    Let $\mX$ be a sub-Gaussian data matrix, and $\vy = \mX^{\T}\vw^* + \vb$ where $\vb$ is the additive and possibly adversarial corruption. Then there exists a gradient-descent algorithm such that $\norm{\vw^{(t)} - \vw^*}_{2} \leq \varepsilon$ after $t = O\left(\log\left(\frac{1}{\sqrt{n}}\frac{\norm{\vb}_{2}}{\varepsilon}\right)\right)$ iterations.
\end{theorem}
The algorithm referred to in \Cref{thm:bhatia-thm-5} uses gradient-descent based iterative thresholding, exhibits a logarithmic dependence on $\norm{\vb}_2$, and is applicable to the {\em realizable} setting, i.e. there must be no stochastic noise in $\vy$. More recently, \cite{awasthi:2022} studied the iterative trimmed maximum likelihood estimator. In their algorithm, at each step they find $\vw^{*}$ which maximizes the likelihood of the samples in the trimmed set. 
\begin{theorem}[Theorem 4.2 in \cite{awasthi:2022}]\label{thm:awasthi-thm-4.2}
    Let $\ermX = \braces{(\vx_{i},y_{i})}_{i=1}^{n}$ be the data generated by a Gaussian regression model defined as $y_{i} = \vw^{*}\cdot\vx_{i} + \eta_{i}$ where $\eta_{i} \sim \gN(0,\nu^2)$ and $\vx_{i}$ are sampled from a sub-Gaussian distribution with second-moment matrix $\mI$. Suppose the dataset has $\epsilon$-fraction of label corruption and $n = \Omega\paren*{\frac{d + \log(1/\delta)}{\epsilon^2}}$. Then there exists an algorithm that returns $\widehat\vw$ such that with probability $1-\delta$, 
    \begin{equation*}
        \norm{\widehat\vw - \vw^{*}}_{2} = O\paren*{\nu \epsilon \log(1/\epsilon)}.
    \end{equation*}
\end{theorem}
Our first result recovers \Cref{thm:awasthi-thm-4.2} as a special case and also allows for vector targets. Furthermore, we obtain the approximation in time $O\paren*{\poly(N,d)\log(R/\epsilon)}$, improving upon the $O(\polylog(N,d)/\epsilon^2)$ runtime complexity of \citep{awasthi:2022}.

\subsection{Contributions}
Our main contribution consists of algorithms and corresponding approximation bounds for gradient-based iterative thresholding for several non-linear learning problems (\Cref{alg:randomized-iterative-thresholding}) and for multitarget linear regression (\Cref{alg:deterministic-linear-regression}). Our proof techniques extend \citep{bhatia2015robust,pmlr-v97-shen19e,awasthi:2022}, as we suppose the adversary also corrupts the covariates. To our knowledge, we are the first to provide guarantees on the performance of iterative thresholding for learning non-linear models, outside of generalized linear models~\citep{awasthi:2022}, under the strong contamination model. 

\begin{table}[!htb]
\centering
\caption{Summary of related work on iterative thresholding algorithms for linear regression under the strong $\epsilon$-contamination model, and our contributions. We assume the target data is sampled from a centered sub-Gaussian distribution with second-moment matrix $\mI$, sub-Gaussian proxy $\mGamma \precsim
 \mI$, and dimension $d$. We assume the variance of the optimal estimator is $\nu$.}
\begin{tabular}{|c|c|c|c|}\hline
Reference & Approximation & Runtime & Algorithm\\\hline
\cite{bhatia2015robust} & $O\paren{\nu}$ & $O\paren*{Nd^{2}\log\paren*{\frac{1}{\sqrt{n}}\frac{\norm{\vb}_{2}}{\epsilon}}}$ & Full Solve \\\hline
\cite{pmlr-v97-shen19e} & $O\paren{\nu}$ & $O\paren*{Nd^{2}\log\paren*{\frac{\norm{\vw^{*}}_{2}}{\nu}}}$ & Gradient Descent \\\hline
\cite{awasthi:2022} & $O\paren{\nu\epsilon\log(1/\epsilon)}$ & $O\paren*{\paren*{Nd^2 + d^3}\paren*{\frac{1}{\nu \epsilon^{2}}}}$ & Full Solve \\\hline
\Cref{corollary:linear-regression} & $O\paren{\nu\epsilon\log(1/\epsilon)}$ & $O\paren*{Nd^2\log\paren*{\frac{\norm{\vw^*}}{\nu\epsilon}}}$ & Gradient Descent \\\hline
\end{tabular}
\label{tab:summary}
\end{table}

\Cref{tab:summary} summarizes the approximation and runtime guarantees for linear regression provided in this work and in the literature. Comparing to \cite{bhatia2015robust}, our runtime does not depend on the $\ell_{2}$ norm of the label corruption, which can be made arbitrarily large by the adversary. We extend upon \cite{pmlr-v97-shen19e} by putting dependence on $\epsilon$ with a small logarithmic factor, improving the error bound significantly. We also offer a significant run-time improvement over the study in \cite{awasthi:2022}, from $O(1/\varepsilon^2)$ to $O(\log(R/\varepsilon))$, where $\norm{\vw^{*}}_{2} \leq R$. 

\begin{table}[!htb]
\centering
\caption{Our results and Distributional assumptions for learning different neuron activation functions. In the approximation bounds, we assume $\vx \sim \gD$ where $\gD$ is a sub-Gaussian distribution with second-moment matrix $\mSigma$ and sub-Gaussian proxy $\mGamma \precsim \mSigma$. We define $\kappa(\mSigma) = \lambda_{\max}(\mSigma)/\lambda_{\min}(\mSigma)$ where $\E_{\vx \sim \gD}\bracks{\vx\vx^{\T}} = \mSigma$, for all $\vx \sim \gD$ and $t \in [T]$ we have $\sigmoid'(\vw^{(t)} \cdot \vx) \geq \gamma$ and $\sigmoid'(\vw^{*} \cdot \vx) \geq \gamma$, for any $x,y \in \sR$ it holds that $\sigmoid(x - y) \leq \norm{\sigmoid}_{\lip}\abs{x - y}$, and $\nu$ is the variance of the optimal estimator. }
\begin{tabular}{|c|c|c|c|}\hline
Reference & Approximation & Neuron & Covariate Distribution\\\hline
\Cref{thm:one-layer-linear-nn-error} & $O\paren*{\nu\epsilon\log(1/\epsilon)}$ &  Linear & Sub-Gaussian \\\hline
\Cref{thm:sigmoid-neuron-approximation-bound} & $O\paren*{\gamma^{-2}\norm{\sigmoid}_{\lip}^{2}\nu\kappa(\mSigma)\sqrt{\epsilon\log(1/\epsilon)}}$ & Sigmoid & Bounded Sub-Gaussian\\\hline
\Cref{thm:leaky-relu-neuron-error} & $O\paren*{\gamma^{-2}\nu\kappa(\mSigma)\sqrt{\epsilon\log(1/\epsilon)}}$& Leaky-\relu & Sub-Gaussian \\\hline
\Cref{thm:relu-neuron-error} & $O\paren*{\nu\kappa(\mSigma)\sqrt{\epsilon\log(1/\epsilon)}}$& \relu & $L_4-L_{2}$ Hypercontractive  \\ \hline
\end{tabular}
\label{tab:distributions}
\end{table}

\Cref{tab:distributions} summarizes our results on learning single neuron models. The bounded assumption in \Cref{thm:sigmoid-neuron-approximation-bound} is necessary to give a lower bound on $\sigmoid'(\vw \cdot \vx)$ for $\vx \sim \gD$ and $\vw$ is equal to $\vw^{(t)}$ for any $t \in [T]$ or $\vw^{*}$. In \Cref{thm:relu-neuron-error}, the $L_{4} \to L_{2}$ hypercontractive assumption allows us to bound the minimum eigenvalue of the sample covariance matrix in the intersection of halfspacse defined by $\vx\vx^{\T}\cdot\vone\braces{\vw^{*}\cdot\vx \geq 0}\cdot\vone\braces{\vw^{(t)}\cdot\vx \geq 0}$ for any $t \in [T]$. In general, our nonlinear learning algorithms have approximation error on the order of $O(\sqrt{\epsilon\log(1/\epsilon))}$. 

All our main results allow corruption to be present in both the covariates and the labels. We are able to show with only knowledge of the minimum and maximum eigenvalues of the sample covariance matrix, iterative thresholding is capable of directly handling corrupted covariates.  In comparison, the algorithm of~\citet{awasthi:2022} considers corruption only in the labels; they extend it to handle corruption in the covariates by preprocessing the covariates using the near-linear filtering algorithm of \citet{dong2019quantum}. 

\textbf{Paper Outline:} In \Cref{sec:preliminaries}, we give the mathematical notation used in the main body of the paper before discussing the literature on iterative thresholding for robust learning. In \Cref{sec:convergence}, we present our formal results and a proof sketch for the learning of linear and non-linear neurons. We defer all proofs of our main results to \Cref{sec:linear-regression-proof,sec:nonlinear-neuron-proofs}. 
\section{Preliminaries} \label{sec:preliminaries}
\subsection{Mathematical Notation and Background}
{\bf Notation.} We use $[T]$ to denote the set $\{1,2,\ldots, T\}$. We say $y \lesssim x$ or $y = O(x)$ if there exists a constant $C$ such that $y \leq Cx$. We say $y \gtrsim x$ or $y = \Omega(x)$ if there exists a constant $C$ s.t. $y \geq Cx$. We define $\sS^{d-1}$ as the $d-1$-dimensional sphere $\{\vx \in \sR^{d}: \norm{\vx}_2 = 1\}$. We denote the Hadamard product between two vectors of the same size as $\vx \circ \vy$; that is, $(\vx \circ \vy)_{i} = x_{i} y_{i}$. 

{\bf Matrices.} For a matrix $\mA$, let $\lambda_{\max}(\mA)$ and $\lambda_{\min}(\mA)$ represent the maximum and minimimum eigenvalues of $\mA$, respectively. Let $\sigma_i(\mA)$ denote the $i$th largest singular values of $\mA$; as such, $\sigma_1(\mA) \geq \sigma_2(\mA) \geq \cdots \geq \sigma_{m \land n}(\mA)$. 
The trace of a square $n \times n$ matrix is given by $\Tr(\mA) = \sum_{i \in [n]}\sigma_{i}(\mA)$. We use the following matrix norms for a matrix $\mA \in \sR^{m \times n}$: 
\begin{align}
    &\text{Spectral Norm: }\norm{\mA}_{2} = \max_{\vx \in \sS^{n-1}}\norm{\mA\vx}_2 = \sigma_{1}(\mA) \\
    &\text{Frobenius Norm: }\norm{\mA}_{\F}^{2} = \Tr(\mA^{\T}\mA) = \sum_{i \in [m \land n]}\sigma_{i}^{2}(\mA).
\end{align}

\noindent\textbf{Probability.} We now discuss the probabilitistic concepts used in this work. We consider the general sub-Gaussian design, which is prevalent in the study of robust statistics (see e.g. \citep{awasthi:2022,bhatia2015robust,Jambulapati2020}). 
\begin{definition}[Sub-Gaussian Distribution] \label{defn:sub-gaussian}
    We say a vector $\vx$ is sampled from a sub-Gaussian distribution with second-moment matrix $\mSigma$ and sub-Gaussian proxy $\mGamma$ if, for any $\vv \in \sS^{d-1}$,
    \begin{equation*}
        \E_{\vx \sim \gD}\bracks*{\exp\paren*{t\vx\cdot \vv}} \leq \exp\paren*{\frac{t^2\norm{\mGamma}_{2}}{2}}\;\forall t \in \sR.
    \end{equation*}
    A scalar random variable $X$ is sub-Gaussian if there exists $K > 0$ such that for all $p \in \sN$,  
    \begin{equation*}
        \norm{X}_{L_{p}} \overset{\defn}{=} \paren*{\E\abs{X}^{p}}^{1/p} \leq K \sqrt{p}.
    \end{equation*}
\end{definition}
Sub-Gaussian distributions are convenient to work with in robust statistics as the empirical mean of any subset of a set of i.i.d realizations of sub-Gaussian random variables is close to the mean of the distribution (see e.g. \cite{steinhardt:2018}). Sub-Gaussian distributions have tails that decay exponentially, i.e. at least as fast as Gaussian random variables. In \Cref{tab:distributions} we introduce $L_{4} \to L_{2}$ hypercontractivity, which we will formally define here. 
\begin{definition}
    A distribution $\gD$ is $L_{4} \to L_{2}$ hypercontractive if there exists a constant $L$ such that 
    \begin{equation*}
        \E_{\vx \sim \gD}\norm{\vx}_{2}^{4} \leq L \E_{\vx \sim \gD}\norm{\vx}_{2}^{2} = L\Tr(\mSigma).
    \end{equation*}
\end{definition}
We defer more technical preliminaries to the relevant proofs in the Appendix. 
\subsection{Related Work}
Iterative thresholding for robust statistical estimation dates back to 1806 by Legendre \cite{legendre1806nouvelles}. Iterative thresholding has been studied theoretically and applied empirically to various machine learning problems including linear regression, GLMs, and generative adversarial networks (GANs) \citep{bhatia2015robust,hu2023outlier,mukhoty:2019}. However, theoretical guarantees for its efficacy in learning nonlinear models are sparse and not sufficiently strong to justify practical usage of iterative thresholding. 

We first introduce the statistical idea of a breakdown point. The breakdown point is defined as the smallest fraction of observations that can be replaced with arbitrary values to cause an estimator to take on arbitrarily large incorrect values. The algorithms presented in this paper have breakdown point $\Omega\paren{1}$. This is an improvement over the robust algorithm given in \citet{chen2013robust} which has breakdown point $\Omega\paren{1/\sqrt{d}}$. Recent papers on iterative thresholding (see e.g. \citep{bhatia2015robust,pmlr-v97-shen19e,awasthi:2022} also have breakdown point $\Omega\paren{1}$.

\citet{bhatia2015robust} study iterative thresholding for least squares regression / sparse recovery. In particular, one of their contributions is a gradient descent algorithm, \textsc{Torrent-GD}, applicable when the covariates are sampled from a sub-Gaussian distribution. Their approximation bound ( \Cref{thm:bhatia-thm-5}) relies on the fact that $\lambda_{\min}(\mSigma) = \lambda_{\max}(\mSigma)$, so that with sufficiently large sample size and sufficiently small corruption parameter $\epsilon$, the condition number $\kappa(\mX)$ approaches 1. \citet{bhatia2015robust} also provide guarantees on the performance of a full solve algorithm, \textsc{Torrent-FC}, which after each thresholding step to obtain $(1-\epsilon)N$ samples sets $\vw^{(t)}$ to be the minimizer of the squared loss over the selected $(1-\epsilon)N$ points. They study this algorithm in the presence of both adversarial and intrinsic noise. Their analysis guarantees $O(\nu)$ error when the intrinsic noise is sub-Gaussian with sub-Gaussian norm $O(\nu)$. 

\citet{pmlr-v97-shen19e}~study iterative thresholding for learning generalized linear models (GLMs). In both the linear and non-linear case, their algorithms exhibit linear convergence. Their results imply a bound of $O(\nu)$ in the linear case. They further provide experimental evidence of the success of iterative thresholding when applied to neural networks. 

More recently,~\citet{awasthi:2022} studied the iterative trimmed maximum likelihood estimator for General Linear Models. Similar to \textsc{Torrent-FC}, their algorithm solves the MLE problem over the data kept after each thresholding step. They prove the best known bounds for iterative thresholding algorithms in the linear regression case, $O(\nu\epsilon\log(1/\epsilon))$. The algorithm studied by \citet{awasthi:2022} natively handles corruptions in labels, and to handle the case of corrupted variates, they first run a near-linear filtering algorithm from \citet{dong2019quantum} to obtain covariates that are sub-Gaussian with close to identity covariance.
\section{Iterative Thresholding for gradient-based learning}
\label{sec:convergence}
In this section we introduce our algorithms for iterative thresholding gradient-based robust learning of linear and non-linear models. We start with the simple case of regression with multiple targets, as this provides a simple introduction to and instantiation of our general proof technique. As a corollary, we find a result regarding linear regression in the sub-Gaussian setting without covariate corruption; this result is compared with the existing literature \citep{awasthi:2022,bhatia:2017,pmlr-v97-shen19e}. Next, we consider the learning of non-linear neurons. This results in a suite of novel results regarding the use of iterative thresholding gradient-based learning that are incomparable with the existing literature.

\subsection{Warm-up: Multivariate Linear Regression}
We will first present our results for the well-studied problem of linear regression in the Huber-$\epsilon$ contamination model. Our results will extend the results in \citet[Theorem 5]{bhatia:2017} and \citet[Lemma A.1]{awasthi:2022} by including covariate corruption without requiring a filtering algorithm, allowing variance in the optimal estimator, and accommodating a second-moment matrix for the uncorrupted data that is not the identity. The loss function for the multivariate linear regression problem for $\mW \in \sR^{K \times d}$, $\mX \in \sR^{d \times n}$, and $\mY \in \sR^{K \times n}$ is 
\begin{equation*}
    \gL(\mW;\mX, \mY) = \norm{\mW\mX - \mY}_{\F}^{2}.
\end{equation*}
We furthermore define $\gR(\mW;\mX,\mY) = \frac{1}{(1-\epsilon)N}\gL(\mW;\mX,\mY)$ as the empirical risk function. We will first give some notation prior to presenting the algorithm. 
\begin{definition}[Hard Thresholding Operator]\label{defn:hard-thresholding}
    For any vector $\vx \in \sR^{n}$, define the order statistics as $\paren{\vx}_{(1)} \leq \paren{\vx}_{(2)} \ldots \leq \paren{\vx}_{(n)}$, the hard thresholding operator is given as
    \begin{equation*}
        \HT(\vx; k) = \braces*{i \in [n]: \paren{\vx}_{i} \in \braces*{\paren{\vx}_{(1)}, \ldots, \paren{\vx}_{(k)}}}.
    \end{equation*}
\end{definition}
With the hard thresholding operator from \Cref{defn:hard-thresholding} in hand, we are now ready to present our algorithm for multi-linear regression. 
\begin{algorithm}
	\caption{Deterministic Gradient Descent Iterative Thresholding for Multi-Linear Regression}
	\textbf{input:} Possibly corrupted $\mX \in \sR^{d \times N}$ with outputs $\mY \in \sR^{K \times N}$, corruption parameter $\epsilon = O(1)$, and $\norm{\vx_{i}} \leq B$ for all $i \in \ermQ$ \\
	\textbf{output:} $O\paren{\nu\epsilon\sqrt{BK\log(1/\epsilon)}}-$Approximate solution $\mW \in \sR^{K \times d}$ to minimize $\norm{\mW - \mW^{*}}_{\F}$
    \begin{algorithmic}[1]
        \State $\mW^{(0)} \gets 0$
        \State $\eta \gets 0.1 \kappa(\mSigma)$
        \State $T \gets O\paren*{\kappa^2(\mSigma)\log\paren*{\frac{\norm{\mW^{*}}_{\F}}{\varepsilon}}}$
		\For{$t \in \bracks{T}$}\label{line:thresh}
            \State $\zeta_{i}^{(t)} = \norm{\mW^{(t)}\vx_{i} - \vy_{i}}^{2}$ \; $\forall i \in [N]$ \Comment{Calculate $\gL$ for each sample}
            \State $\ermS^{(t)} \gets \HT(\vzeta^{(t)}, (1-\epsilon)N)$ \Comment{See \Cref{defn:hard-thresholding}}
            \State $\mW^{(t+1)} \gets \mW^{(t)} - \eta \nabla \gR(\mW^{(t)};\ermS^{(t)})$ \Comment{Gradient Descent Update}
        \EndFor
	\end{algorithmic}
    \textbf{return:} $\mW^{(T)}$
    \label{alg:deterministic-linear-regression}
\end{algorithm}

\noindent \textbf{Runtime. }In each iteration we calculate the $\ell_{2}$ error for $N$ points, in total $O(Nd)$. For the Hard Thresholding step, it suffices to find the $n(1-\epsilon)$-th largest element, we can run a selection algorithm in worst-case time $O(N\log N)$, then partition the data in $O(N)$. The run-time for calculating the gradient and updating $\vw^{(t)}$ is dominated by the matrix multiplication in $\mX_{\ermS^{(t)}}\mX_{\ermS^{(t)}}^{\T}$ which can be done in $O(Nd^2)$. Then considering the choice of $T$, we have the algorithm runs in time $O\left(Nd^2\log\left(\frac{\norm{\vw^{*}}_{2}}{\nu\epsilon}\right)\right)$ to obtain $O(\nu \epsilon\log(1/\epsilon))$ $\ell_{2}$-approximation error. 
\begin{theorem}\label{thm:one-layer-linear-nn-error}
Let $\mX = \bracks{\vx_{1},\ldots,\vx_{n}}^{\T} \in \sR^{d \times N}$ be the data matrix and $\mY = \bracks{\vy_{1},\ldots,\vy_{n}} \in \sR^{K \times N}$ be the output, such that for $i \in \ermP$, $\vx_{i}$ are sampled from a sub-Gaussian distribution second-moment matrix $\mSigma$ and sub-Gaussian proxy $\mGamma$, and $\norm{\vx_{j}} \leq B$ for $j \in \ermQ$. Suppose $\vy_{i} = \mW^{*}\vx_{i} + \ve_{i}$ where $\ve_{i} \sim \gN(\vzero, \sigma^{2}\mI)$ for all $i \in \ermP$. Then after $O\paren{\kappa(\mSigma)\log\paren{\norm{\mW^{*}}_{\F}/\varepsilon}}$ gradient descent iterations, $N = \Omega\paren*{d/\epsilon^{2}}$, and learning rate $\eta = 0.1 \lambda_{\max}^{-2}(\mSigma)$, with probability exceeding $1-3Te^{-\Omega(d)}$, \Cref{alg:deterministic-linear-regression} returns $\mW^{(T)}$ such that
\begin{equation*}
	\norm{\mW^{(T)} - \mW^*}_{\F} \leq \varepsilon + O\paren*{\nu \epsilon  \sqrt{KB\log(1/\epsilon)}}.
\end{equation*}
\end{theorem}
We are able to recover the result of Lemma 4.2 in \cite{awasthi:2022} when the covariates (corrupted and un-corrupted) are sampled from a sub-Gaussian distribution with second-moment matrix $\mI$ and $\mGamma \precsim \mI$. The full solve algorithm studied in \cite{awasthi:2022} returns a $O\paren*{\nu\epsilon \log(1/\epsilon)}$ in time $O\paren*{\frac{1}{\epsilon^{2}}\paren*{Nd^2 + d^3}}$ and the same algorithm studied in \cite{bhatia2015robust}, \textsc{Torrent-FC}  otains $O(\nu)$ approximation error in run-time $O\paren*{\log\paren*{\frac{1}{\sqrt{n}}\frac{\norm{\vb}}{\nu\epsilon\log(1/\epsilon)}}\paren{Nd^2 + d^3}}$, with the gradient descent based approach, we are able to improve the runtime to $O\paren*{\log\paren*{\frac{\norm{\vw^{*}}}{\nu\epsilon}}Nd^2}$ for the same approximation bound. By no longer requiring the full-solve, we are able to remove super-linear relation to $d$. In comparison to \cite{bhatia2015robust}, we do not have dependence on the noise vector $\vb$, which can have very large norm in relation to the norm of $\vw^{*}$. Our proof is also a significant improvement over the presentation given in Lemma 5 of \cite{pmlr-v97-shen19e} as under the same conditions, we give more than the linear convergence, but we show linear convergence is possible on any second-moment matrix of the good covariates and covariate corruption, and then develop concentration inequality bounds to match the best known result for iteratived trimmed estimators. We will formalize our results into a corollary to give a more representative comparison in the literature. 
\begin{corollary}\label{corollary:linear-regression}
Let $\mX = \bracks{\vx_{1},\ldots,\vx_{n}}^{\T} \in \sR^{d \times N}$ be the data matrix and $\vy = \bracks{y_{1},\ldots,y_{n}} \in \sR^{N}$ be the output, such that $\vx_{i}$ are sampled from a sub-Gaussian distribution with second-moment matrix $\mI$ and sub-Gaussian proxy $\mI$ and $\vy_{i} = \vw^{*}\cdot \vx_{i} + \xi_{i}$ where $\xi_{i} \sim \gN(0, \nu^2)$ for $i \in \ermP$. Then after $O\left(\log\left(\frac{\norm{\vw^{*}}_{2}}{\varepsilon}\right)\right)$ gradient descent iterations, sample size $N = \Omega\paren*{\frac{d + \log(1/\delta)}{\epsilon}}$, and learning rate $\eta = 0.1$, with probability exceeding $1-\delta$, \Cref{alg:deterministic-linear-regression} returns $\vw^{(T)}$ such that
\begin{equation*}
	\norm{\vw^{(T)} - \vw^{*}}_{2} \leq \varepsilon + O\paren*{\nu \epsilon  \sqrt{KB\log(1/\epsilon)}}.
\end{equation*}    
Suppose for $i \in Q$, $\vx_{i}$ are sampled from a sub-Gaussian distribution with sub-Gaussian Norm $K$ and second-moment matrix $\mI$. Then, 
\begin{equation*}
    \norm{\vw^{(T)} - \vw^{*}}_{2} \leq \varepsilon + O\paren*{\nu \epsilon\log(1/\epsilon)}.
\end{equation*}
\end{corollary}
The second relation given in \Cref{corollary:linear-regression} matches the best known bound for robust linear regression with iterative thresholding. The first relation given in \Cref{corollary:linear-regression} is the extension to handle second-moment matrices which do not have unitary condition number as well as corrupted covariates. 
\subsection{Activation Functions}
We first give properties of the non-linear functions we will be learning. All functions we henceforth study will have some subset of the below listed properties. 
\begin{property} \label{property:monotonic}
    $\sigmoid$ is a continuous, monotically increasing, and differentiable almost everywhere. 
\end{property}
\begin{property} \label{property:lipschitz}
    $\sigmoid$ is Lipschitz, i.e. $\abs{\sigmoid(x) - \sigmoid(y)} \leq \norm{\sigmoid}_{\lip} \abs{x - y}$.  
\end{property}
\begin{property} \label{property:lower-bound}
    For any $x \geq 0$, there exists $ \gamma > 0$ such that $\inf_{\abs{z} \leq x}\sigmoid'(z) \geq \gamma > 0$. 
\end{property}
Sigmoid functions such as $\tanh$ and sigmoid and the leaky-$\relu$ function satisfy Properties~\ref{property:monotonic},~\ref{property:lipschitz}, and~\ref{property:lower-bound}. \Cref{property:lower-bound} does not hold for the $\relu$ function, and therefore we require stronger conditions for our approximation bounds to hold.
\subsection{Learning Sigmoidal Neurons}
We now study the problem of minimizing the $\ell_{2}$ loss for sigmoidal type neurons. For a single training sample $(\vx_{i},y_{i}) \sim \gD$, the loss is given as follows,  
\begin{equation*}
	\gL\left(\vw;\vx_i,y_i\right) = (\sigmoid(\vw \cdot \vx_{i}) - y_i)^2.
\end{equation*}
\begin{algorithm}
	\caption{Gradient Descent Iterative Thresholding for Learning a Non-linear Neuron}
	\textbf{input:} Possibly corrupted $\mX \in \sR^{d \times N}$ with outputs $\vy \in \sR^{N}$, activation function $\nu$, corruption parameter $\epsilon = O(1)$, and small constant $\alpha$. \\
	\textbf{output:} $\nu \sqrt{\epsilon\log(1/\epsilon)}-$Approximate solution $\vw \in \sR^{d}$ to minimize $\norm{\vw - \vw^{*}}_{2}$. 
    \begin{algorithmic}[1]
        \State $\vw^{(0)} \sim \gB_{d}(\alpha \norm{\vw^{*}})$
        \State $\eta \gets 0.1 \kappa^{-2}(\mSigma)$
        \State $T \gets O\paren*{\kappa^2(\mSigma)\log\paren*{\frac{\norm{\vw^{*}}_{2}}{\varepsilon}}}$
		\For{$t \in \bracks{T}$}
            \State $\zeta_{i}^{(t)} = (\sigmoid(\vx_{i}^{\T}\vw^{(t)}) - y_i)^2$ \; $\forall i \in [N]$ \Comment{Calculate $\gL$ for each point}
            \State $\ermS^{(t)} \gets \HT(\vzeta^{(t)}, (1-\epsilon)N)$ \Comment{See \Cref{defn:hard-thresholding}}
            \State $\vw^{(t+1)} \gets \vw^{(t)} - \eta \nabla \gR(\vw^{(t)};\ermS^{(t)})$ \Comment{Gradient Descent Update}
        \EndFor
	\end{algorithmic}
    \textbf{return:} $\vw^{(T)}$
    \label{alg:randomized-iterative-thresholding}
\end{algorithm}
\begin{theorem}\label{thm:sigmoid-neuron-approximation-bound}
Let $\mX = \bracks{\vx_{1},\ldots,\vx_{N}}^{\T} \in \sR^{d \times N}$ be the data matrix and $\vy = \bracks{y_{1},\ldots,y_{n}}$ be the output, such that $\vx_{i}$ are sampled from a sub-Gaussian distribution with second-moment matrix $\mSigma$ and sub-Gaussian proxy $\mGamma$, and $y_{i} = \sigmoid(\vw^{*}\cdot\vx_{i}) + \xi_{i}$ for $\xi_{i} \sim \gN(0, \nu^{2})$ for all $i \in \ermP$. Suppose the activation function, satisfies Properties~\ref{property:monotonic},~\ref{property:lipschitz}, and~\ref{property:lower-bound}. Then after $O\left(\kappa(\mSigma)\log\left(R/\varepsilon\right)\right)$ (where $\norm{\vw^{*}}_{2} \leq R$) gradient descent iterations, then with probability exceeding $1-\delta$,
\begin{equation*}
	\norm{\vw^{(T)} - \vw^{*}}_{2} \leq O\paren*{\gamma^{-2}\lambda_{\min}^{-1}(\mSigma)\norm{\sigmoid}_{\lip}^{2}\epsilon\sqrt{B\log(1/\epsilon)}} + O\paren*{\gamma^{-2}\norm{\sigmoid}_{\lip}^{2}\nu\kappa(\mSigma)\sqrt{\epsilon\log(1/\epsilon)}}.
\end{equation*}
when $N = \Omega\paren*{\frac{d + \log(1/\delta)}{\epsilon^{2}}}$.
\end{theorem}
The result is slightly weaker than the approximation bound achieved in \Cref{thm:one-layer-linear-nn-error} by a square root factor. In the proof for our linear regression result, we are able to leverage the consistency of linear regression estimators, i.e. as $N \to \infty$, we have $\argmin_{\vw}\gL(\vw, \gD) \to \argmin_{\vw}\E_{\gD}\bracks*{\gL(\vw, \gD)}$. 
\begin{proof}[Sketch]\label{sec:proof-sketch}
    We will give a general sketch of the proof for our sigmoidal neuron result, deferring the more technical ideas to the proof in the appendix. For any $t \in [T]$, we have 
    \begin{align}
        \norm{\vw^{(t+1)} - \vw^{*}} &\leq \norm{\vw^{(t)} - \eta\nabla\gR(\vw^{(t)};\ermS^{(t)})  - \vw^{*}}_{2} \nonumber \\
        &\leq  \underbrace{\norm{\vw^{(t)} - \vw^{*} - \eta\nabla\gR(\vw^{(t)};\ermS^{(t)} \cap \ermP)}}_{I} + \underbrace{\norm{\eta \nabla \gR(\vw^{(t)};\ermS^{(t)} \cap \ermQ)}_{2}}_{II}. \label{eq:sketch-expansion}
    \end{align}
    In the above, the second relation follows from the linearity of the loss function, which gives
    \begin{equation*}
        \nabla \gR(\vw^{(t)};\ermS^{(t)}) = \nabla \gR(\vw^{(t)};\ermS^{(t)} \cap \ermP) + \nabla \gR(\vw^{(t)};\ermS^{(t)} \cap \ermQ),
    \end{equation*}
    and then applying the triangle inequality. We upper bound $I$ through its square,
    \begin{equation*}
        I^2 = \norm{\vw^{(t)} - \vw^{*}}_{2}^{2} - \underbrace{2 \eta \cdot \angle{\vw^{(t)} - \vw^{*}, \nabla \gR(\vw^{(t)};\ermS^{(t)} \cap \ermP)}}_{I_{1}} + \underbrace{\eta^2 \cdot \norm{\nabla \gR(\vw^{(t)}; \ermS^{(t)} \cap \ermP)}^{2}}_{I_{2}}.
    \end{equation*}
    In this step the finer details of the proof for leaky-$\relu$ and $\relu$ will differ, however the structure remains the same. We will prove there exists a constant $c_{1} > 0$ such that
    \begin{equation*}
        I_{1} \geq (1-2\epsilon) c_{1} \norm{\vw^{(t)} - \vw^{*}}_{2}^{2} - c_{3}\norm{\vw^{(t)} - \vw^{*}}_{2},
    \end{equation*}
    where $c_{3}$ is a term that is dependent on the variance of the Gaussian noise, one of our contributions is that $c_{3} = O(\nu\sqrt{\epsilon\log(1/\epsilon)})$ when $N$ sufficiently large for the activation functions studied in the text. Next, an application of Peter-Paul's inequality\footnote{Peter-Paul's inequality states that for any $p,q > 1$ such that $\frac{1}{p} + \frac{1}{q} = 1$, then for every $t$, we have $ab \leq \frac{t^{p}a^{p}}{p} + \frac{t^{-q}b^{q}}{q}$. Consider Young's Inequality and replace $a$ with $at^{p}$ and $b$ with $bt^{-p}$. } gives us, 
    \begin{equation*}
        I_{1} \geq \paren*{(1-2\epsilon)c_{1} - c_{3}c_{1}}\norm{\vw^{(t)} - \vw^{*}}_{2}^{2}  - c_{3}c_{1}^{-1}.
    \end{equation*}
    We next show there exists a positive constant $c_{2}$, such that 
    \begin{equation*}
        I_{2}^{2} \leq (1-\epsilon)c_{2} \norm{\vw^{(t)} - \vw^{*}}_{2}^{2}.
    \end{equation*}
    Then, solving a simple quadratic equation, we have that $\eta^2 C_2 \leq \eta c_{1} c_{3} $ for a $c_{3} \in (0,1)$ when we choose $\eta \leq \frac{c_{1}c_{3}}{c_{2}}$ and we are able to eliminate the norm of the gradient squared term. We must now control the corrupted gradient term. The key idea is to note that from the optimality of the sub-quantile set, 
    \begin{equation}\label{eq:key-proof-step}
        \sum_{i \in \ermS^{(t)} \cap \ermQ} \gL(\vw^{(t)};\vx_{i},y_{i}) \leq \sum_{i \in \ermP \setminus \ermS^{(t)}} \gL(\vw^{(t)};\vx_{i},y_{i})
    \end{equation}
    and $\abs{\ermS^{(t)} \cap \ermQ} = \abs{\ermP \setminus \ermS^{(t)}}$. We then prove the existence of a constant $c_{4}$ such that,
    \begin{equation*}
        II \leq \epsilon c_{4} \norm{\vw^{(t)} - \vw^{*}}_{2}^{2} 
    \end{equation*}
    Then, combining our results, we end up with a linear convergence of the form, 
    \begin{equation*}
        \norm{\vw^{(t+1)} - \vw^{*}}_{2} \leq \norm{\vw^{(t)} - \vw^{*}}_{2} \paren*{1 - \eta (1-2\epsilon)c_{1} + \eta \epsilon c_{4}} + c_{3}c_{1}^{-1}
    \end{equation*}
    We obtain a bound that is of the form, 
    \begin{equation*}
        \norm{\vw^{(t+1)} - \vw^{*}}_{2} \leq \norm{\vw^{(t)} - \vw^{*}}_{2}\paren*{1 - \lambda} + E
    \end{equation*}
    Then, we find that 
    \begin{equation*}
        \norm{\vw^{(t)} - \vw^{*}}_{2} \leq \norm{\vw^{(0)} - \vw^{*}}_{2} \paren*{1 - \lambda}^{t} + \sum_{k \in [t]}(1-\lambda)^{k} E
    \end{equation*}
    We then find asymptotically, the second term converges to $\lambda^{-1}E$. Then, it suffices to find $T$ such that, 
    \begin{equation*}
        \norm{\vw^{(T)} - \vw^{*}}_{2} \leq \lambda^{-1}E
    \end{equation*}
    We can note that $1 - \lambda \leq e^{-\lambda}$, then bounding $T$, we obtain a $\lambda^{-1}E$ approximation bound when
    \begin{equation*}
        T \geq \lambda^{-1}\cdot \log\paren*{\frac{\norm{\vw^{*} - \vw^{(0)}}_{2}}{\lambda^{-1}E}}
    \end{equation*}
    In our deterministic algorithm, we choose $\vw^{(0)} = \vzero$ and thus we have $\norm{\vw^{*} - \vw^{(0)}}_{2} = \norm{\vw^{*}}_{2}$. In our randomized algorithm, we have from \Cref{lem:random-initialization}, with high probability $\norm{\vw^{*} - \vw^{(0)}} \leq \norm{\vw^{*}}$, giving us the desired bound. 
\end{proof}
\subsubsection{Algorithmic $\epsilon$}
In practice, one does not have access to the true corruption rate in the dataset. We therefore differentiate between the algorithmic corruption parameter $\epsilon$ and the true corruption parameter of the dataset $\epsilon^{*}$. Consider the case when $\epsilon \geq \epsilon^{*}$ i.e., we overestimate the corruption rate of the dataset. We have at any iteration $t \in [T]$, that $\abs{\ermS_{\epsilon} \cap \ermQ} \geq \abs{\ermS_{\epsilon^{*}} \cap \ermQ}$ as $(1-\epsilon)N \geq (1-\epsilon^{*})N$. We similarly have $\abs{\ermP \setminus \ermS^{(t)}_{\epsilon^{*}}} \leq \abs{\ermP \setminus \ermS_{\epsilon}^{(t)}}$ as the thresholded set is smaller in cardinality. We thus have our key step, \Cref{eq:key-proof-step}, will still hold when $\epsilon \geq \epsilon^{*}$ and we can thus obtain the same approximation bounds with only a lower bound on $\epsilon^{*}$. 

\subsection{Learning Leaky-ReLU Neurons}\label{sec:nn}
We will now consider learning a neuron with the Leaky-ReLU function. We can note that \Cref{property:monotonic},~\ref{property:lipschitz}, and~\ref{property:lower-bound} all hold for the Leaky-ReLU. More conveniently, we have $\gamma$ in \Cref{property:lower-bound} is constant over $\sR$. In our proof we are able to leverage the fact that the second derivative is zero almost surely. 
\begin{theorem}\label{thm:leaky-relu-neuron-error}
    Let $\mX = \bracks{\vx_{1},\ldots,\vx_{n}}^{\T} \in \sR^{N \times d}$ be the data matrix and $\vy = \bracks{y_{1}\ldots,y_{n}}^{\T}$ be the output. Suppose $\vx_{i}$ are sampled from a sub-Gaussian distributionsecond-moment matrix $\mSigma$ and proxy $\mGamma$, and $y_i = \sigmoid(\vw^{*} \cdot \vx_{i}) + \xi_{i}$ for $\xi_{i} \sim \gN(0,\nu^2)$ where $\sigmoid(x) = \max\braces{\gamma x, x}$ for all $i \in \ermP$. Then after $O\left(\gamma^{-2}\kappa\paren*{\mSigma}\log\left(\frac{\norm{\vw^{*}}}{\varepsilon}\right)\right)$  gradient descent iterations and $\epsilon \leq \frac{\gamma^{2}\lambda_{\min}(\mSigma)}{\sqrt{32B\lambda_{\max}(\mSigma)}}$, with probability exceeding $1-4\delta$, \Cref{alg:randomized-iterative-thresholding} with learning rate $\eta = O\paren{\kappa^{-2}(\mSigma)}$ returns $\vw^{(T)}$ such that 
    \begin{equation*}
        \norm{\vw^{(T)} - \vw^*}_{2} \leq O\paren*{\nu\norm{\mGamma}_{2}\lambda_{\min}^{-1}(\mSigma)\sqrt{\epsilon\log(1/\epsilon)}} + O\paren*{\gamma^{-2}\kappa(\mSigma)\nu\epsilon\sqrt{B\log(1/\epsilon)}}.
    \end{equation*}
\end{theorem}
\begin{proof}
    The proof is deferred to \Cref{sec:leaky-relu-neuron-error}. 
\end{proof}
Our proof and result also hold for a smoothened version of the Leaky-\relu. To avoid the kink at $x = 0$, one can consider the Smooth-Leaky-$\relu$, defined as  
\begin{equation*}
    \text{Smooth-Leaky-\relu}(x) = \alpha x + (1-\alpha) \log\paren*{1 + e^x}
\end{equation*}
for an $\alpha \in (0,1)$. The Smooth-Leaky-\relu\; satisfies Properties~\ref{property:monotonic},~\ref{property:lipschitz}, and~\ref{property:lower-bound}, and is convex, indicating that \Cref{thm:leaky-relu-neuron-error} can be applied. 

\subsection{Learning ReLU Neurons}
We will now consider the problem of learning ReLU neural networks. We first give a preliminary result for randomized initialization. 
\begin{lemma}[Theorem 3.4 in \cite{du2018gradient}]\label{lem:random-initialization}
    Suppose $\vw^{(0)}$ is sampled uniformly from a $p$-dimensional ball with radius $\alpha\norm{\vw^{*}}$ such that $\alpha \leq \sqrt{\frac{1}{2\pi p}}$, then with probability at least $\frac{1}{2} - \alpha\sqrt{\frac{\pi p}{2}}$,
    \begin{equation*}
        \norm{\vw^{(0)} - \vw^{*}}_{2} \leq \sqrt{1 - \alpha^2}\norm{\vw^{*}}_{2}.
    \end{equation*}
\end{lemma}
From this result we are able to derive probabistic guarantees on the convergence of learning a $\relu$ neuron. 
\begin{theorem}\label{thm:relu-neuron-error}
    Let $\mX = \bracks{\vx_{1},\ldots,\vx_{n}}^{\T} \in \sR^{n \times d}$ be the data matrix and $\vy = \bracks{y_{1},\ldots, y_{n}}^{\T}$ be the output, such that for $\vx_{i}$ are sampled from a sub-Gaussian distribution with second-moment matrix $\mSigma$ and sub-Gaussian proxy $\mGamma$ and the output is given as $y_{i} = \sigmoid(\vw^{*} \cdot \vx_{i}) + \xi_{i}$ for $\xi_{i} \sim \gN(0,\nu^2)$ for all $i \in \ermP$. Then after $O\paren*{\kappa(\mSigma)\log\paren*{\frac{\norm{\vw^{*}}}{\varepsilon}}}$ gradient descent iterations and $N = \Omega\paren*{\frac{d + \log(1/\delta)}{\epsilon}}$, then with probability exceeding $1-3T\delta$, \Cref{alg:randomized-iterative-thresholding} with learning rate $\eta = O\paren*{\frac{\lambda_{\min}(\mSigma)}{\lambda_{\max}^{2}(\mSigma)}}$ returns $\vw^{(T)}$ such that
    \begin{equation*}
        \norm{\vw^{(T)} - \vw^*}_{2} \leq O\paren*{\nu\norm{\mGamma}_{2}\lambda_{\min}^{-1}(\mSigma)\sqrt{\epsilon\log(1/\epsilon)}} + O\paren*{\kappa(\mSigma)\nu\epsilon\sqrt{B\log(1/\epsilon)}}.
    \end{equation*}
\end{theorem}
\begin{proof}
    The proof is deferred to \Cref{sec:relu-neuron-error}.
\end{proof}
Our proof for learning $\relu$ neurons follows the same high level structure as learning sigmoidal or leaky-$\relu$ neurons, however it is significantly more technical, as we require bounds on the spectra of empirical covariance matrices over the intersection of half-spaces. We also note that \Cref{lem:random-initialization} implies that randomized restarts with high probability will return a vector with $O\paren{\sqrt{\epsilon\log(1/\epsilon)}}$ $\ell_{2}$ approximation error. With access to an uncorrupted test set, using concentration of measure inequalities for sub-Gaussian distributions, it is possible to use adaptive algorithms to determine how many randomized restarts is enough to obtain $O\paren{\sqrt{\epsilon\log(1/\epsilon)}}$ approximation error. 
	
\section{Discussion}
In this paper, we study the theoretical convergence properties of iterative thresholding for non-linear learning problems in the Strong $\epsilon$-contamination model. Our warm-up result for linear regression reduces the runtime while achieving the best known approximation for iterative thresholding algorithms. Many papers have experimentally studied the iterative thresholding estimator in large scale neural networks \citep{hu2023outlier,pmlr-v97-shen19e} and to our knowledge, we are the first paper to make advancements in the theory of iterative thresholding for a general class of activation functions. There are many directions for future work. Regarding iterative thresholding, our paper has established upper bounds on the approximation error of activation functions, an interesting next step is on upper bounds for the sum of activation functions, i.e. one hidden-layer neural networks. In the linear regression case, \citet{gao20robust} derived the minimax optimal error of $O(\sigma\epsilon)$. Establishing this result for sigmoidal, leaky-\relu, and \relu\; functions would be helpful in the discussing the strength of our bounds. Deriving upper and lower bounds for iterative thresholding for binary classification is a good direction for future research. In $\pm 1$ classification, considering $y = \sign(\vw^{*}\cdot \vx + \xi)$, the $\sign$ function adds adds an interesting complication. A study on if our current techniques can also handle the $\sign$ function would be interesting. 

\bibliographystyle{plainnat}
\bibliography{robust-nonlinear-learning.bib}
\clearpage
	
\appendix
\section{Proofs for Linear Regression} \label{sec:linear-regression-proof}

\textbf{Notation.} We first state some notational preliminaries. Partition the indices of the training data set $\ermX$ into the indices of the inliers, $\ermP$, and the indices of the corrupted points, $\ermQ$. That is, let $[N] = \ermP \cup \ermQ$ where $|\ermP| = (1-\epsilon)N$ and $|\ermQ| = \epsilon N$, where $i \in \ermP$ iff $(\vx_{i},y_i)$ is an inlier, and $i \in \ermQ$ iff $(\vx_{j},y_{j})$ has been corrupted by the adversary. For $t \in [T]$, we denote $\ermS^{(t)}$ as the set of indices selected after thresholding at iteration $t$. We decompose $\ermS^{(t)}$ into the indices of {\em True Positives} (inliers) and the indices of {\em False Positives} (corrupted samples), i.e. $\ermS^{(t)} = \left(\ermS^{(t)} \cap \ermP\right) \cup \left(\ermS^{(t)} \cap \ermQ \right) = \TP \cup \FP$. We similarly decompose the points discarded at each iteration into {\em False Negatives} and {\em True Negatives}, i.e. $\ermX \setminus \ermS^{(t)} = \left( \paren{\ermX \setminus \ermS^{(t)}} \cap \ermP \right) \cup \left( \paren{\ermX \setminus \ermS^{(t)}} \cap \ermQ \right)= \FN \cup \TN$.

\subsection{Proof of \Cref{thm:one-layer-linear-nn-error}}\label{sec:one-layer-linear-nn-error}

We first state some useful lemmata. Let $\vect: \sR^{n \times k} \to \sR^{nk}$ represent the vectorizaton of a matrix to a vector placing its columns one by one into a vector, and let $\otimes: \sR^{N \times K} \times \sR^{L \times M} \to \sR^{NL \times KM}$ represent the Kronecker product between two matrices.
We then have the useful facts,
\begin{lemma}\label{lem:vec-trace-inner-product}
    Suppose $\mA, \mB \in \sR^{m \times n}$, then 
    \begin{equation*}
        \angle{\mA, \mB}_{\Tr} \overset{\defn}{=} \Tr(\mA \mB^T)  = \angle{\vect(\mA), \vect(\mB)}.
    \end{equation*}
\end{lemma}

\begin{lemma}\label{lem:vec-kronecker-relation}
    Suppose $\mA, \mB, \mC$ are conformal matrices, then 
    \begin{equation*}
        \vect\paren*{\mA\mB\mC} = \paren{\mC^{\T} \otimes \mA}\vect\paren{\mB}.
    \end{equation*}
\end{lemma}
\begin{proof}
    Recall that $\mX \in \sR^{d \times N}$ and $\mY \in \sR^{K \times N}$ hold the training data. Given any set of indices $\ermS$, the risk of a model $\mW \in \sR^{K \times d}$ on the points in $\ermS$, and the gradient of that risk, are 
    \begin{equation*}
            \gR(\mW;\ermS) = \frac{1}{|\ermS|} \norm{\mW\mX_{\ermS} - \mY_{\ermS}}_{\F}^2.
    \quad \text{and} \quad
        \nabla\gR(\mW; \ermS) = \frac{2}{|\ermS|}(\mW\mX - \mY)\mX^{\T}.
    \end{equation*}
   
   We first show we can obtain Linear Convergence plus a noise term and a consistency term, which we define as the difference between the optimal estimator in expectation and the optimal estimator over a population. We then show the consistency term goes to zero with high probability as $N \to \infty$. Finally, we use the concentration inequalities developed in \Cref{sec:probability-theory} to give a clean approximation bound.  \\
   \textbf{Step 1: Linear Convergence}. We will first show iterative thresholding has linear convergence to optimal. Let $\widehat\mW = \argmin_{\mW \in \sR^{K \times d}}\gR(\mW;\ermP)$ be the minimizer over the good data. Then, we have 
    \begin{align*}
        &\norm{\mW^{(t+1)} - \mW^*}_{\F} = \norm{\mW^{(t)} - \mW^* - \eta\nabla\gR(\mW^{(t)};\ermS^{(t)})}_{\F}\\
        &= \norm{\mW^{(t)} - \mW^* - \eta\nabla\gR(\mW^{(t)};\TP) + \eta\nabla\gR(\mW^{*};\ermP) - \eta\nabla\gR(\mW^{*};\ermP) + \eta\nabla\gR(\widehat\mW;\ermP) - \eta\nabla\gR(\mW^{(t)};\FP)}_{\F} \\
        &\leq \underbrace{\norm{\mW^{(t)} - \mW^* - \eta\nabla\gR(\mW^{(t)};\TP) + \eta\nabla\gR(\mW^{*};\TP)}_{\F}}_{I} + \underbrace{\norm{\eta\nabla\gR(\mW^{(t)};\FP)}_{\F}}_{II} + \underbrace{\norm{\eta\nabla\gR(\mW^{*};\FN)}_{\F}}_{III} \nonumber \\
        &\quad + \underbrace{\norm{\eta\nabla\gR(\mW^{*};\ermP) - \eta\nabla\gR(\widehat\mW;\ermP)}_{\F}}_{IV}. 
    \end{align*}
    In contrast with our proof sketch in \Cref{sec:proof-sketch}, we introduce a consistency estimate in $IV$. We first will upper bound $I$ through the expansion of its square.
    \begin{multline*}
        \norm{\mW^{(t)} - \mW^* - \eta\nabla\gR(\mW^{(t)};\TP) + \eta\nabla \gR(\mW^{*};\TP)}_{\F}^{2} = \norm{\mW^{(t)} - \mW^*}_{\F}^2\\
        - \underbrace{2\eta \cdot \Tr\bigl((\mW^{(t)} - \mW^{*})^{\T}(\nabla\gR(\mW^{(t)};\TP) - \nabla\gR(\mW^{*};\TP))\bigr)}_{I_{1}} + \underbrace{\norm{\eta\nabla\gR(\mW^{(t)};\TP) - \eta\nabla \gR(\mW^{(t)};\TP)}_{\F}^2}_{I_{2}} .
    \end{multline*}
    We first lower bound $I_{1}$, we will obtain a lower bound that is less than $I_{2}$, allowing us to cancel the $I_{2}$ term.  
    \begin{equation*}
        \begin{aligned}
            I_{1} &= 2\eta \cdot \Tr\bigl((\mW^{(t)} - \mW^{*})^{\T}(\nabla\gR(\mW^{(t)};\TP) - \nabla\gR(\mW^{*};\TP))\bigr) \\
            \overset{\defn}&{=} \frac{4\eta}{\paren{1-\epsilon}N} \cdot\Tr\bigl((\mW^{(t)} - \mW^{*})^{\T}\bigl((\mW^{(t)}\mX_{\TP} - \mY_{\TP})\mX_{\TP}^{\T} - \mE_{\TP}\mX_{\TP}^{\T}\bigr)\bigr) \\
            \overset{(i)}&{=} \frac{4\eta}{(1-\epsilon)N}\cdot \Tr\paren*{(\mW^{(t)} - \mW^*)^{\T}(\mW^{(t)} - \mW^{*})\mX_{\TP}\mX_{\TP}^{\T}}.
        \end{aligned}
    \end{equation*}
    In the above, $(i)$ follows from recalling that $\mY_{\TP} = \mW^{*}\mX_{\TP} - \mE_{\TP}$. We then have, 
    \begin{equation*}
        \begin{aligned}
            I_{1} \overset{(ii)}&{=} \frac{4\eta}{(1-\epsilon)N}\cdot \Tr\paren*{(\mW^{(t)} - \mW^{*})\mX_{\TP}\mX_{\TP}^{\T}\paren{\mW^{(t)} - \mW^{*}}^{\T}} \\
            \overset{(iii)}&{=} \frac{4\eta}{(1-\epsilon)N}\cdot \angle{\vect(\paren{\mW^{(t)} - \mW^{*}}^{\T}), \vect\paren{\mX_{\TP}\mX_{\TP}^{\T}(\mW^{(t)} - \mW^{*})^{\T}}} \\
            \overset{(iv)}&{=} \frac{4\eta}{(1-\epsilon)N}\cdot \angle{\vect(\paren{\mW^{(t)} - \mW^{*}}^{\T}), \paren{\mI \otimes \mX_{\TP}\mX_{\TP}^{\T}}\vect((\mW^{(t)} - \mW^{*})^{\T})} \\
            \overset{(v)}&{=} \frac{4\eta}{(1-\epsilon)N}\cdot\sum_{k \in [K]}\bigl\langle \vw_{k}^{(t)} - \vw_{k}^*, \mX_{\TP}\mX_{\TP}^{\T}\paren{\vw_{k}^{(t)} - \vw_{k}^{*}}\bigr\rangle \\
            \overset{(vi)}&{\geq} \frac{2\eta}{(1-\epsilon)N}\cdot\sum_{k \in [K]} \bigl( \lambda_{\min}(\mX_{\TP}\mX_{\TP}^{\T})\norm{\vw_{k}^{(t)} - \vw_{k}^{*}}_{2}^{2} + \norm{\mX_{\TP}\mX_{\TP}^{\T}}_{2}^{-1}\norm{\mX_{\TP}\mX_{\TP}^{\T}\paren{\vw_{k}^{(t)} - \vw_{k}^{*}}}_{2}^{2} \bigr) \\
            &= \frac{2\eta}{(1-\epsilon)N}\cdot\lambda_{\min}(\mX_{\TP}\mX_{\TP}^{\T})\norm{\mW^{(t)} - \mW^*}_{\F}^{2} + \underbrace{\frac{2\eta}{(1-\epsilon)N}\cdot\norm{\mX_{\TP}\mX_{\TP}^{\T}}_{2}^{-1}\norm{(\mW^{(t)} - \mW^*)\mX_{\TP}\mX_{\TP}^{\T}}_{\F}^{2}}_{I_{12}}.
        \end{aligned}
    \end{equation*}
    In the above, $(ii)$ follows from the cyclic property of the trace, $(iii)$ follows from the relation given in \Cref{lem:vec-trace-inner-product}, $(iv)$ holds from the relation given in  \Cref{lem:vec-kronecker-relation}, the inequality in $(vi)$ follows from \Cref{lem:bubeck-3.11}, and in $(v)$ we apply \Cref{lem:vec-kronecker-relation}, which gives the following equality,
    \begin{equation*}
        \paren{\mI \otimes \mX_{\TP}\mX_{\TP}^{\T}}\vect\paren{\paren{\mW^{(t)} - \mW^{*}}^{\T}} = 
        \begin{bmatrix}
            \mX_{\TP}\mX_{\TP}^{\T} & & \\
            & \ddots & \\
            & & \mX_{\TP}\mX_{\TP}^{\T}
        \end{bmatrix}
        \begin{bmatrix}
            \vw_{1} - \vw_{1}^{*}\\
            \vdots \\
            \vw_{K} - \vw_{K}^{*}
        \end{bmatrix}
        = 
        \begin{bmatrix}
            \mX_{\TP}\mX_{\TP}^{\T}\paren{\vw_{1} - \vw_{1}^{*}} \\
            \vdots \\
            \mX_{\TP}\mX_{\TP}^{\T}\paren{\vw_{K} - \vw_{K}^{*}}
        \end{bmatrix}.
    \end{equation*}
    We now upper bound the corrupted gradient term.
    \begin{equation*}
        \begin{aligned}
            II \overset{\defn}&{=} \frac{2\eta}{(1-\epsilon)N}\cdot \norm{(\mW^{(t)}\mX_{\FP} - \mY_{\FP})\mX_{\FP}^{\T}}_{\F} \\
            \overset{(vii)}&{\leq} \frac{2\eta}{(1-\epsilon)N}\cdot\norm{\mX_{\FP}}_{2}\norm{\mW^{(t)}\mX_{\FP} - \mY_{\FP}}_{\F} \\
            \overset{(viii)}&{\leq} \frac{2\eta}{(1-\epsilon)N}\cdot\norm{\mX_{\FP}}_{2}\norm{\mW^{(t)}\mX_{\FN} - \mY_{\FN}}_{\F} \\
            \overset{(ix)}&{\leq} \frac{2\eta}{(1-\epsilon)N}\cdot \norm{\mX_{\FP}}_{2}\bigl(\norm{\mW^{(t)}\mX_{\FN} - \mW^*\mX_{\FN}}_{\F}  + \norm{\mE_{\FN}}_{\F}\bigr)\\
            \overset{(x)}&{\leq} \frac{2\eta}{(1-\epsilon)N}\cdot \norm{\mX_{\FP}}_{2}\norm{\mX_{\FN}}_{2}\norm{\mW^{(t)} - \mW^*}_{\F} + \frac{2\eta}{(1-\epsilon)N}\cdot \norm{\mX_{\FP}}_{2}\norm{\mE_{\FN}}_{\F}.
        \end{aligned}
    \end{equation*}
    In the above, the equalities in $(vii)$ and $(x)$ from the fact that for any two size compatible matrices, $\mA, \mB$, it holds that $\norms{\mA\mB}_{\F} \leq \norms{\mA}_{\F}\norms{\mB}_{2}$, $(viii)$ follows from the optimality of the Hard Thresholding Operator (see \Cref{defn:hard-thresholding}), and $(ix)$ follows from the sub-additivity of the Frobenius norm. We will now upper bound $III$. 
    \begin{equation*}
        III = \norm{\eta\nabla\gR(\mW^{*};\FN)}_{\F} \overset{\defn}{=} \frac{2\eta}{(1-\epsilon)N}\cdot\norm{(\mW^{*}\mX_{\FN} - \mY_{\FN})\mX_{\FN}^{\T}}_{\F}  \leq \frac{2\eta}{(1-\epsilon)N}\cdot \norm{\mE_{\FN}\mX_{\FN}^{\T}}_{\F}.
    \end{equation*}
    In the above, we use the fact that for any two conformal matrices, $\mA, \mB$, it holds that $\norm{\mA\mB}_{\F} \leq \norm{\mA}_{\F}\norm{\mB}_{2}$. \\
    \textbf{Step 2: Consistency}. We will now upper bound the consistency estimate, $IV$. 
    \begin{equation*}
        \begin{aligned}
            IV \overset{\defn}&{=} \norm{\eta\nabla\gR(\mW^{*};\ermP) - \eta\nabla\gR(\widehat\mW;\ermP)}_{\F} \\
            &= \frac{2\eta}{(1-\epsilon)N} \cdot \norm{\paren{\widehat\mW - \mW^{*}}\mX_{\ermP}\mX_{\ermP}^{\T}}_{\F} \leq \frac{2\eta}{(1-\epsilon)N}\cdot \norm{\mE_{\ermP}\mX_{\ermP}^{\T}}_{\F}.
        \end{aligned}
    \end{equation*}
    We can then have from \Cref{lem:max-scaled-gaussian-matrix-norm}, 
    \begin{equation*}
        \norm{\mE_{\ermP}\mX_{\ermP}^{\T}}_{\F}^{2} \leq 2\sigma^{2}\paren*{K\norm{\mX_{\ermP}}_{\F}^{2}\log\paren*{2N^2/\delta}} \leq 2\sigma^{2}\paren*{(1-\epsilon)Nd\lambda_{\max}(\mSigma)\log\paren*{2N^2/\delta}}.
    \end{equation*}
    We then have with failure probability at most $\delta$, 
    \begin{equation*}
       \frac{2\eta}{(1-\epsilon)N}\cdot \norm{\mE_{\ermP}\mX_{\ermP}^{\T}}_{\F} \leq \eta \cdot \sqrt{\frac{8\sigma^{2}d\lambda_{\max}(\mSigma)\log(2N^2/\delta)}{(1-\epsilon)N}}.
    \end{equation*}
    We then have from our choice of $\eta = 0.1 \lambda^{-1}_{\max}(\mSigma)$, we have $I_{2} \leq I_{12}$ with high probability. Then, from noting that $\sqrt{1-2x} \leq 1-x$ for any $x \leq 1/2$, we obtain
    \begin{multline*}
        \norm{\mW^{(t+1)} - \mW^*}_{\F} \leq \norm{\mW^{(t)} - \mW^{*}}_{\F}\paren*{1 - \frac{2\eta}{(1-\epsilon)N}\cdot \lambda_{\min}(\mX_{\TP}\mX_{\TP}^{\T}) + \frac{2\eta}{(1-\epsilon)N}\cdot\norm{\mX_{\FP}}_{2}\norm{\mX_{\FN}}_{2}} \\
        + \frac{2\eta}{(1-\epsilon)N}\cdot \norm{\mX_{\FP}}_{2}\norm{\mE_{\FN}}_{\F} + \frac{2\eta}{(1-\epsilon)N}\cdot \norm{\mE_{\FN}\mX_{\FN}^{\T}}_{\F} + \eta \cdot \sqrt{\frac{32\sigma^{2}d\lambda_{\max}(\mSigma)\log(2N^2/\delta)}{3(1-\epsilon)N}}.
    \end{multline*}
    \textbf{Step 3: Concentration Bounds.} From \Cref{prop:chi-squared-sum} and our $\ell_{2}$ bounded corrupted covariate assumption, we obtain with failure probability at most $\delta$, 
    \begin{equation*}
        \norm{\mE_{\FN}}_{\F}\norm{\mX_{\FP}}_{\F} \leq N\epsilon \sigma \sqrt{30KB\log(1/\epsilon)}.
    \end{equation*}
    From \Cref{lem:max-scaled-gaussian-matrix-norm}, we have when $N \geq \log(1/\delta)$, with failure probability at most $\delta$,
    \begin{equation*}
        \norm{\mE_{\FN}\mX_{\FN}^{\T}}_{\F} \leq \sqrt{6K\log N }\norm{\mX_{\FN}}_{2}.
    \end{equation*}
    Then utilizing the result from \Cref{lem:sub-gaussian-matrix-eigenvalues}, we have with failure probability at most $\delta$, 
    \begin{equation*}
        \sqrt{6K\log N }\norm{\mX_{\FN}}_{2} \leq \sqrt{60K\lambda_{\max}(\mSigma) N\log N \cdot \epsilon \log(1/\epsilon)}.
    \end{equation*}
    Noting that $\abs{\ermS^{(t)} \cap P} \geq (1-2\epsilon)N$, we have from \Cref{lem:sub-gaussian-matrix-eigenvalues} for $\epsilon \leq \frac{1}{60}\cdot\kappa^{-1}(\mSigma)$, the minimum eigenvalue satisfies with failure probability at most $\delta$, 
    \begin{equation*}
        \lambda_{\min}(\mX_{\TP}\mX_{\TP}^{\T}) \geq \frac{N}{4}\cdot \lambda_{\min}(\mSigma) .
    \end{equation*}
    Then when $\epsilon \leq \sqrt{\frac{1}{960B}\cdot \kappa^{-1}(\mSigma) \lambda_{\min}(\mSigma)}$, we have with high probability, 
    \begin{equation*}
        \norm{\mX_{\FP}}_{2}\norm{\mX_{\FN}}_{2} - \lambda_{\min}\paren{\mX_{\TP}\mX_{\TP}^{\T}} \leq \frac{\eta}{4(1-\epsilon)}\cdot \lambda_{\min}(\mSigma).
    \end{equation*}
    Combining our estimates, we have 
    \begin{multline*}
        \norm{\mW^{(t+1)} - \mW^{*}}_{\F} \leq \norm{\mW^{(t)} - \mW^{*}}_{\F} \paren*{1 - \frac{\eta}{4(1-\epsilon)}\cdot \lambda_{\min}(\mSigma)} + \eta \cdot \sigma \epsilon\sqrt{480KB\log(1/\epsilon)} \\
        + \frac{\eta}{(1-\epsilon)\sqrt{N}}\cdot \sqrt{240K\log(N)\lambda_{\max}(\mSigma)\epsilon\log(1/\epsilon)} + \frac{\eta}{(1-\epsilon)\sqrt{N}}\cdot \sigma \sqrt{11d\lambda_{\max}(\mSigma)\log(2N^2/\delta)}.
    \end{multline*}
    with probability exceeding $1 - 3\delta$. Then, when $N = \widetilde\Omega\paren*{\frac{d\lambda_{\max}(\mSigma) + \log(1/\delta)}{\epsilon^2KB}}$ we have
    \begin{equation*}
        \norm{\mW^{(t+1)} - \mW^{*}}_{\F} \leq \norm{\mW^{(t)} - \mW^{*}}_{\F} \paren*{1 - \frac{\eta}{4(1-\epsilon)}\cdot \lambda_{\min}(\mSigma)} + \eta \cdot \sigma \epsilon\sqrt{4320KB\log(1/\epsilon)}.
    \end{equation*}
    Then, solving for the induction with an infinite sum, referring to the proof sketch in \Cref{sec:proof-sketch} and our choice of $\eta$, we have after $O\paren*{\kappa\paren*{\mSigma}\cdot\log\paren*{\frac{\norm{\mW^{*}}_{\F}}{\varepsilon}}}$ iterations, 
    \begin{equation*}
        \norm{\mW^{(T)} - \mW^{*}}_{\F} \leq \varepsilon + \frac{\sigma\epsilon\sqrt{34560KB\log(1/\epsilon)}}{\lambda_{\min}(\mSigma)}.
    \end{equation*}
     Our proof is complete. 
\end{proof}

\section{Proofs for Learning Nonlinear Neurons}\label{sec:nonlinear-neuron-proofs}
In this section we present our omitted proofs for the approximation bounds for learning nonlinear neurons with \Cref{alg:randomized-iterative-thresholding}. 
\subsection{Sigmoidal Neurons}
In this section we will give the ommited proofs for the $\ell_2$ approximation bounds for learning sigmoidal neurons with \Cref{alg:randomized-iterative-thresholding} in the Strong $\epsilon$-Contamination Model. 
\subsubsection{Proof of \Cref{thm:sigmoid-neuron-approximation-bound}}\label{sec:sigmoid-neuron-approximation-bound}
\begin{proof}
    From \Cref{alg:randomized-iterative-thresholding}, we have the gradient update for learning a Sigmoid neuron for the $\ell_{2}$ loss.
    \begin{equation*}
        \vw^{(t+1)} = \vw^{(t)} - \frac{2\eta}{(1-\epsilon)N}\cdot\sum_{i \in S^{(t)}} (\sigmoid(\vw^{(t)}\cdot\vx_{i}) - y_i)\cdot \sigmoid'(\vw^{(t)} \cdot \vx_{i}) \cdot \vx_{i}.
    \end{equation*}
    Our proof will follow a similar structure to the proof for linear regression. We first show we can obtain linear convergence of $\vw^{(t)}$ to $\vw^{*}$ with some error. Then we rigorously analyze the concentration inequalities to give crisp bounds on the upper bound for $\epsilon$ and show the noise term is $O(\epsilon \log(1/\epsilon))$. 
    \\\textbf{Step 1: Linear Convergence.} 
    \begin{equation*}
        \begin{aligned}
            \norm{\vw^{(t+1)} - \vw^{*}}_{2} &= \norm{\vw^{(t)} - \eta\nabla\gR(\vw^{(t)};\ermS^{(t)}) - \vw^{*}}_{2}\\
            &= \norm{\vw^{(t)} - \vw^{*} - \eta\nabla \gR(\vw^{(t)};\TP) - \eta\nabla\gR(\vw^{(t)};\FP)}_{2} \\
            &\leq \underbrace{\norm{\vw^{(t)} - \vw^{*} - \eta\nabla \gR(\vw^{(t)};\TP)}_{2}}_{I} + \underbrace{\norm{\eta\nabla\gR(\vw^{(t)};\FP)}_{2}}_{II} .
        \end{aligned}
    \end{equation*}
    We will expand $I$ through its square and give an upper bound.  
    \begin{equation*}
        I^{2} = \norm{\vw^{(t)} - \vw^{*}}_{2}^{2} - \underbrace{2\eta \cdot \angle{\vw^{(t)} - \vw^{*}, \nabla\gR(\vw^{(t)};\TP)}}_{I_1} + \underbrace{\eta^2\cdot \norm{\nabla\gR(\vw^{(t)};\TP)}_{2}^{2}}_{I_2} . 
    \end{equation*}
    We now lower bound $I_{1}$. Note from the randomized initialization, we have $\norm{\vw^{(0)} - \vw^{*}} \leq \vw^{*}$. Then, noting that $\norm{\vw^{*}} \leq R$, we have by the Cauchy-Schwarz inequality, for any $\vx \sim \gP$, we have $\abs{\vw^{(t)} \cdot \vx} \leq 2RB$ almost surely. Here we leverage \Cref{property:lower-bound} and note there exists a $\gamma$ s.t. $\sigmoid'(x) \geq \gamma > 0$ for all $x \in \sR$ s.t. $x \leq 2RB$, from which we obtain
    \begin{equation*}\begin{aligned}
        I_{1} &= \frac{4\eta}{(1-\epsilon)N}\cdot \angle{\vw^{(t)} - \vw^{*},\sum_{i \in \TP}\paren{\sigmoid(\vw^{(t)}\cdot\vx_{i}) - y_i}\cdot \sigmoid'(\vw^{(t)}\cdot\vx_{i})\cdot \vx_{i}} \\
        &= \frac{4\eta}{(1-\epsilon)N}\cdot \angle{\vw^{(t)} - \vw^{*},\sum_{i \in \TP}\paren{\sigmoid(\vw^{(t)}\cdot\vx_{i}) - \sigmoid(\vw^{*}\cdot\vx_{i}) + \xi_{i}}\cdot \sigmoid'(\vw^{(t)}\cdot\vx_{i})\cdot \vx_{i}} \\
        \overset{(i)}&{\geq} \underbrace{\frac{4\eta}{(1-\epsilon)N}\cdot \gamma^{2}\lambda_{\min}\paren{\mX_{\TP}\mX_{\TP}^{\T}} \norm{\vw^{(t)} - \vw^{*}}_{2}^{2}}_{I_{11}} - \frac{4\eta}{(1-\epsilon)N}\cdot \norm{\vw^{(t)} - \vw^{*}}_{2} \normsm{ \sum_{i \in \TP} \xi_{i}\sigmoid'(\vw^{(t)}\cdot\vx_{i})\cdot \vx_{i}}_{2}.
    \end{aligned}\end{equation*}
    In the above, $(i)$ follows from defining the function $h: \sR^{d} \mapsto \sR$,
    \begin{equation*}
        \vw \mapsto \int_{\sR^{d}}\paren*{\int \sigmoid}\paren{\vw\cdot \vx} dP(\vx),
    \end{equation*}
    where $dP(\vx) = \vone\braces{\vx \in \TP}$. Then from \Cref{property:lower-bound}, we have that $\sigmoid'$ is strictly positive over a compact domain, this implies that $\int\sigmoid$ is strongly convex over a compact domain. Then we can calculate the Hessian, 
    \begin{equation*}
        \nabla^2 h(\vw) = \int_{\sR^{d}}\sigmoid'(\vw\cdot \vx) \cdot \vx\vx^{\T} dP(\vx) \succeq \gamma \lambda_{\min}\paren{\mX_{\TP}\mX_{\TP}^{\T}} \cdot \mI.
    \end{equation*}
    With the strong convexity in hand, we have, 
    \begin{multline*}
        \angle{\vw^{(t)} - \vw^{*}, \sum_{i \in \TP}\paren{\sigmoid(\vw^{(t)}\cdot \vx_{i}) - \sigmoid(\vw^{*}\cdot\vx_{i}} \cdot \vx_{i}}\\ = \angle{\vw^{(t)} - \vw^{*}, \int_{0}^{1}\nabla^2 h(\vw^{*} + \theta (\vw^{(t)} - \vw^{*})) d\theta \cdot (\vw^{(t)} - \vw^{*})}
        \overset{(ii)}{\geq} \gamma \lambda_{\min}(\mX_{\TP}\mX_{\TP}^{\T})\cdot \norm{\vw^{(t)} - \vw^{*}}_{2}^{2}.
    \end{multline*}
    In the above, $(ii)$ follows from noting that for any $\theta \in [0,1]$ we have, 
    \begin{equation*}
        \norm{(1-\theta)\vw^{*} + \theta \vw^{(t)}}_{2} \leq \theta \norm{\vw^* - \vw^{(t)}}_{2} + \norm{\vw^{*}}_{2} \leq 2\norm{\vw^{*}}_{2} = 2R.
    \end{equation*}
    We can then use the same $\gamma$ as previously defined. 
    Then from an application of Peter-Paul's Inequality, we obtain
    \begin{multline*}
        \frac{4\eta}{(1-\epsilon)N}\cdot \norm{\vw^{(t)} - \vw^{*}}_{2} \normsm{ \sum_{i \in \TP} \xi_{i}\sigmoid'(\vw^{(t)}\cdot\vx_{i})\cdot \vx_{i}}_{2} 
        \leq \frac{\eta}{(1-\epsilon)N}\cdot \gamma^{2}\lambda_{\min}(\mX_{\TP}\mX_{\TP}^{\T})\norm{\vw^{(t)} - \vw^{*}}_{2}^{2}\\
        + \frac{4\eta}{(1-\epsilon)N}\cdot\gamma^{-2}\lambda_{\min}^{-1}(\mX_{\TP}\mX_{\TP}^{\T})\normsm{\sum_{i \in \TP} \xi_{i}\sigmoid'(\vw^{(t)}\cdot\vx_{i})\cdot \vx_{i}}_{2}^{2}.
    \end{multline*}
    We next upper bound $I_{2}$. We note that $\sigmoid$ is $\norm{\sigmoid}_{\lip}$-Lipschitz, then from an application of the triangle inequality, we obtain
    \begin{equation*}\begin{aligned}
        I_{2} \overset{\defn}&{=} \frac{4\eta^2}{\bracks*{(1-\epsilon)N}^{2}}\cdot\normsm{\sum_{i \in \TP}\paren{\sigmoid(\vw^{(t)}\cdot\vx_{i}) - \sigmoid\paren{\vw^{*}\cdot\vx_{i}} + \xi_{i}}\cdot\sigmoid'(\vw^{(t)}\cdot\vx_{i})\cdot \vx_{i}}_{2}^{2} \\
        \overset{(i)}&{=} \frac{4\eta^2}{\bracks*{(1-\epsilon)N}^{2}}\cdot\normsm{\sum_{i \in \TP}\vx_{i}\vx_{i}^{\T}\paren{\vw - \vw^{*}}\cdot \sigmoid'(c_{i}) \sigmoid'(\vw^{(t)} \cdot \vx_{i}) + \xi_{i}\cdot\sigmoid'(\vw^{(t)}\cdot\vx_{i})\cdot \vx_{i}}_{2}^{2}\\
        &\leq \underbrace{\frac{8\eta^{2}}{\bracks*{(1-\epsilon)N}^{2}}\cdot \norm{\sigmoid}_{\lip}^{4}\lambda_{\max}^{2}(\mX_{\TP}\mX_{\TP}^{\T})\norm{\vw^{(t)} - \vw^{*}}_{2}^{2}}_{I_{21}} + \frac{8\eta^{2}}{\bracks*{(1-\epsilon)N}^{2}}\cdot \normsm{\sum_{i \in \TP}\xi_{i}\sigmoid'(\vw^{(t)}\cdot\vx_{i})\cdot \vx_{i}}_{2}^{2}  .
    \end{aligned}\end{equation*}
    In the above, $(i)$ follows from noting there exists a constant $c_{i} \in \bracks{\vw^{(t)}\cdot\vx_{i}, \vw^{*}\cdot\vx_{i}}$ such that 
    \begin{equation*}
        \sigmoid'(c_{i})\paren{\vw^{(t)} - \vw^{*}} \cdot \vx_{i} = \sigmoid(\vw^{(t)}\cdot\vx_{i}) - \sigmoid(\vw^{*}\cdot\vx_{i})
    \end{equation*}
    from the Mean-Value Theorem. Then, from choosing $\eta \leq \frac{\gamma^{2}(1-\epsilon)N\lambda_{\min}(\mX_{\TP}\mX_{\TP}^{\T})}{4\norm{\sigmoid}_{\lip}^{4}\lambda_{\max}^{2}(\mX_{\TP}\mX_{\TP}^{\T})}$. We have $I_{21} \leq 0.5 I_{11}$. We now bound the corrupted gradient term. 
    \begin{equation*}\begin{aligned}
        II^{2} &= \frac{4\eta^2}{\bracks{(1-\epsilon)N}^2}\cdot \normsm{\sum_{i \in \FP}\paren{\sigmoid(\vw^{(t)}\cdot\vx_{i}) - y_i}\cdot \sigmoid'(\vw \cdot \vx_{i})\cdot \vx_{i}}_{2}^2 \\
        \overset{(ii)}&{\leq} \frac{4\eta^2}{\bracks{(1-\epsilon)N}^2}\cdot \norm{\sigmoid}_{\lip}^{2} \norm{\mX_{\FP}\mX_{\FP}^{\T}}_{2}\sum_{i \in \FP}\paren{\sigmoid(\vw^{(t)}\cdot\vx_{i}) - y_i}^{2} \\
        \overset{(iii)}&{\leq}\frac{4\eta^2}{\bracks{(1-\epsilon)N}^2}\cdot \norm{\sigmoid}_{\lip}^{2} \norm{\mX_{\FP}\mX_{\FP}^{\T}}_{2}\sum_{i \in \FN}\paren{\sigmoid(\vw^{(t)}\cdot\vx_{i}) - y_i}^{2} \\
        &= \frac{4\eta^2}{\bracks{(1-\epsilon)N}^2}\cdot \norm{\sigmoid}_{\lip}^{2} \norm{\mX_{\FP}\mX_{\FP}^{\T}}_{2}\sum_{i \in \FN}\paren{\sigmoid(\vw^{(t)}\cdot\vx_{i}) - \sigmoid(\vw^{*}\cdot\vx_{i}) + \xi_{i}}^{2} \\
        \overset{(iv)}&{\leq} \frac{8\eta^2}{\bracks{(1-\epsilon)N}^2}\cdot \norm{\sigmoid}_{\lip}^{2} \norm{\mX_{\FP}\mX_{\FP}^{\T}}_{2}\paren*{\norm{\sigmoid}_{\lip}^{2}\cdot \norm{\mX_{\FN}\mX_{\FN}^{\T}}_{2} \norm{\vw^{(t)} - \vw^{*}}_{2}^{2} + \norm{\vxi_{\FN}}_{2}^{2}}  .
    \end{aligned}\end{equation*}
    In the above, $(ii)$ follows from \Cref{lem:hadamard-vector-matrix}, $(iii)$ follows from the optimality of the Hard-Thresholding operator, $(iv)$ follows from noting $\sigmoid$ is Lipschitz and the elementary inequality $(a+b)^2 \leq 2a^2 + 2b^2$ for any $a,b \in \sR$. Concluding the step, we have,
    \begin{multline*}
        \norm{\vw^{(t+1)} - \vw^{*}}_{2} \leq \norm{\vw^{(t)} - \vw^{*}}_{2}\paren*{1 - \frac{\eta}{2(1-\epsilon)N}\cdot \gamma^{2}\lambda_{\min}(\mX_{\TP}\mX_{\TP}^{\T}) + \frac{\sqrt{8}\eta}{(1-\epsilon)N}\cdot \norm{\sigmoid}_{\lip}^{2}\norm{\mX_{\FP}}_{2}\norm{\mX_{\FN}}_{2}} \\
        + \paren*{\frac{\sqrt{8}\eta}{(1-\epsilon)N} + \frac{2\sqrt{\eta}}{\sqrt{(1-\epsilon)N}}\cdot\lambda_{\min}^{-1/2}(\mX_{\TP}\mX_{\TP}^{\T})}\normsm{\sum_{i \in \TP}\xi_{i}\sigmoid'(\vw^{(t)}\cdot\vx_{i}) \cdot \vx_{i}}_{2} + \frac{\sqrt{8}\eta}{(1-\epsilon)N}\cdot\norm{\mX_{\FP}}_{2}\norm{\vxi_{\FN}}_{2}.
    \end{multline*}
    \textbf{Step 2: Concentration Bounds.} From \Cref{lem:nonlinear-neuron-noise-bound-lipschitz}, we have with probability at least $1-\delta$, 
    \begin{equation*}
        \norm{\sum_{i \in \TP}\xi_{i}\sigmoid'(\vw^{(t)}\cdot\vx_{i})\cdot \vx_{i}}_{2} \lesssim \nu\norm{\mGamma}_{2}\norm{\sigma}_{\lip}N \sqrt{\epsilon\log(1/\epsilon)}.
    \end{equation*}
    Recall that $\norm{\mX_{\FP}}_{2} \leq \sqrt{\epsilon N B}$. From \Cref{lem:sub-gaussian-matrix-eigenvalues}, with probability at least $1 - \delta$, we have
    \begin{equation*}
        \norm{\mX_{\FN}}_{2}\norm{\mX_{\FP}}_{2} \leq N\epsilon\cdot\sqrt{10B \log(1/\epsilon)}.
    \end{equation*}
    From the second relation in \Cref{lem:sub-gaussian-matrix-eigenvalues}, we have when $N = \Omega\paren*{\frac{d + \log(1/\delta)}{\epsilon}}$, with probability exceeding $1-\delta$, the minimimum eigenvalue satisfies, 
    \begin{equation*}
        \lambda_{\min}(\mX_{\TP}\mX_{\TP}^{\T}) \geq \frac{N}{4} \lambda_{\min}(\mSigma).
    \end{equation*}
    We then find for 
    \begin{equation*}
        \epsilon^{2} \leq \frac{\gamma^{2}}{\norm{\sigmoid}_{\lip}^{2}}\frac{\lambda_{\min}(\mSigma)}{\sqrt{B\lambda_{\max}(\mSigma)}}
    \end{equation*}
    and probability exceeding $1-\delta$, 
    \begin{equation*}
        (\gamma^2/2)\lambda_{\min}(\mX_{\TP}\mX_{\TP}^{\T}) - \sqrt{8}\norm{\sigmoid}_{\lip}^{2}\norm{\mX_{\FP}}_{2}\norm{\mX_{\FN}}_{2} \geq \frac{N\gamma^{2}}{16} \lambda_{\min}(\mSigma).
    \end{equation*}
    Combining the $\ell_{2}$ boundedness of the corrupted covariates and \Cref{prop:chi-squared-sum}, we have with probability exceeding $1-\delta$, 
    \begin{equation*}
        \norm{\mX_{\FP}}_{2}\norm{\vxi_{\FN}}_{2} \leq N \epsilon \nu \cdot \sqrt{30B\log(1/\epsilon)}.
    \end{equation*}
    Then combining the results with our choice of $\eta$, we obtain, 
    \begin{equation*}\begin{aligned}
        \norm{\vw^{(t+1)} - \vw^{*}}_{2} &\leq \norm{\vw^{(t)} - \vw^{*}}_{2}\paren*{1 - \frac{\eta}{16(1-\epsilon)N}\cdot \gamma^{2}\lambda_{\min}(\mSigma)} +\eta \epsilon\norm{\sigmoid}_{\lip}^{2}\sqrt{80B\log(1/\epsilon)}\nonumber \\
        &+ \paren*{\sqrt{8}\eta + 2\sqrt{\eta\lambda_{\min}(\mSigma)}}\norm{\sigmoid}_{\lip}^{2}c_{1}N\lambda_{\max}(\mSigma)\nu\sqrt{\epsilon\log(1/\epsilon)}.
    \end{aligned}\end{equation*}
    In the above, $c_{1}$ is a constant and the final inequality holds when $N \geq \frac{Rd \log12 + 2d\log 12 + \log(1/\delta)}{6\epsilon\log(1/\epsilon)} = \Omega\paren*{\frac{d + \log(1/\delta)}{\epsilon}}$. Then, following from our sketch in \Cref{sec:proof-sketch}, we have
    \begin{equation*}
        \norm{\vw^{(t+1)} - \vw^{*}}_{2} \leq \varepsilon + O\paren*{\gamma^{-2}\lambda_{\min}^{-1}(\mSigma)\norm{\sigmoid}_{\lip}^{2}\epsilon\sqrt{B\log(1/\epsilon)}} + O\paren*{\gamma^{-2}\norm{\sigmoid}_{\lip}^{2}\kappa(\mSigma)\sqrt{\epsilon\log(1/\epsilon)}}.
    \end{equation*}
    Our proof is complete. 
\end{proof}
\subsection{Leaky-ReLU Neuron}
In this section we will derive our $\ell_{2}$-approximation bound for the Leaky-$\relu$ neuron.
\subsubsection{Proof of \Cref{thm:leaky-relu-neuron-error}}\label{sec:leaky-relu-neuron-error}
\begin{proof}
    We will decompose the gradient into the good component and corrupted component. The first part of our proof will show that $\vw$ moves in the direction of $\vw^{*}$, then in the second part of the proof we will show the affect of the corrupted gradient. Finally, we combine step 1 and step 2 to show that there exists sufficiently small $\epsilon$ such that we can get linear convergence with a small additive error term. \\
    \textbf{Step 1:} Upper bounding the $\ell_{2}$ norm distance between $\vw^{(t+1)}$ and $\vw^{*}$. We have from Algorithm~\ref{alg:randomized-iterative-thresholding}, 
    \begin{equation*}
        \begin{aligned}
            &\norm{\vw^{(t+1)} - \vw^{*}}_{2} = \norm{\vw^{(t)} - \vw^{*} - \eta\nabla\gR(\vw^{(t)};\ermS^{(t)})}_{2}\\
            &= \norm{\vw^{(t)} - \vw^{*}- \eta\nabla\gR(\vw^{(t)};\TP) + \eta\nabla\gR(\vw^{*};\TP) - \eta\nabla\gR(\vw^{*};\TP) - \nabla\gR(\vw^{(t)};\FP)}_{2} \\
            &\leq \underbrace{\norm{\vw^{(t)} - \vw^{*}- \eta\nabla\gR(\vw^{(t)};\TP) + \eta\nabla\gR(\vw^{*};\TP)}_{2}}_{I} + \underbrace{\norm{\eta\nabla\gR(\vw^{*};\TP)}_{2}}_{II} + \underbrace{\norm{\eta \nabla \gR(\vw^{(t)};\FP)}_{2}}_{III} .
        \end{aligned}
    \end{equation*}
    We will first upper bound the $I_{1}$ by an expansion of its square. 
    \begin{equation*}
        \begin{aligned}
            I^{2}
            &= \norm{\vw^{(t)} - \vw^{*}}_{2}^{2} - \underbrace{2\eta\cdot \angle{\vw^{(t)} - \vw^{*}, \nabla \gR(\vw^{(t)};\TP)} - \nabla\gR(\vw^{(t)};\TP)}_{I_{1}}\\
            &\quad + \underbrace{\eta^{2} \cdot \norm{\nabla \gR\paren{\vw^{(t)};\TP} - \nabla \gR\paren{\vw^{*};\TP}}^{2}}_{I_{2}}.
        \end{aligned}
    \end{equation*}
    In the above, the a.s. relation follows from \Cref{property:monotonic}. 
    We will first lower bound $I_{1}$. We will first bound the spectrum of $\nabla^2 \gL(\vw;\TP)$ for any $\vw \in \sR^{d}$.  
    \begin{equation*}
        \nabla^2\gL(\vw;\TP) = 2\cdot \sum_{i \in \TP}\paren*{\sigmoid(\vw \cdot \vx_{i}) - y_i}\cdot \sigmoid''(\vw \cdot \vx_{i}) \cdot \vx_{i}\vx_{i}^{\T} + 2\cdot \sum_{i \in \TP}\bracks*{\sigmoid'(\vw \cdot \vx_{i})}^{2}\cdot \vx_{i}\vx_{i}^{\T}.
    \end{equation*}
    Then, from noting that the second derivative of Leaky-\relu\; is non-zero at one point, we have
    \begin{equation*}
        \nabla^2\gL(\vw;\TP) \overset{\as}{=} 2 \cdot \sum_{i \in \TP}\bracks*{\sigmoid'(\vw\cdot\vx_{i})}^{2} \cdot \vx_{i} \vx_{i}^{\T} .
    \end{equation*}
    We then obtain for any $\vw \in \sR^{d}$, almost surely, 
    \begin{equation}\label{eq:leaky-relu-hessian-eigenvalues}
        2\cdot \gamma^{2} \lambda_{\min}(\mX_{\TP}\mX_{\TP}^{\T}) \cdot \mI \preceq  \nabla^{2} \gL(\vw;\TP) \preceq 2\cdot \norm{\sigmoid}_{\lip}^{2} \lambda_{\max}(\mX_{\TP}\mX_{\TP}^{\T}) \cdot \mI .
    \end{equation}
    We can now lower bound $I_{1}$ from the convexity of the Leaky-ReLU,
    \begin{equation*}
        \begin{aligned}
            I_{1} 
            &= \frac{4\eta}{(1-\epsilon)N}\cdot \angle{\vw^{(t)} - \vw^{*},\int_{0}^{1}\nabla^2\gR(\vw^{*} + \theta(\vw^{*} -\vw^{(t)});\TP)d\theta \cdot (\vw^{(t)} - \vw^{*})} \\
            \overset{\eqref{eq:leaky-relu-hessian-eigenvalues}}&{\geq} \frac{4\eta}{(1-\epsilon)N}\cdot \gamma^{2}\lambda_{\min}(\mX_{\TP}\mX_{\TP}^{\T})\norm{\vw^{(t)} - \vw^{*}}_{2}^2.
        \end{aligned}
    \end{equation*}
    We now will upper bound $I_{2}$ with a similar argument. 
    \begin{equation*}
        \begin{aligned}
            I_{2} 
            &= \frac{4\eta^2}{\bracks*{(1-\epsilon)N}^2}\cdot\normsm{\int_{0}^{1}\nabla^2\gR(\vw^{*} + \theta(\vw^{*} -\vw^{(t)});\TP)d\theta \cdot (\vw^{(t)} - \vw^{*})}_{2}^{2} \\
            \overset{\eqref{eq:leaky-relu-hessian-eigenvalues}}&{\leq} \frac{4\eta^2}{\bracks*{(1-\epsilon)N}^2}\cdot \norm{\sigmoid}_{\lip}^{2}\lambda_{\max}^{2}(\mX_{\TP}\mX_{\TP}^{\T})\norm{\vw^{(t)} - \vw^{*}}_{2}^{2}.
        \end{aligned}
    \end{equation*}
    where $(ii)$ follows from the $\norm{\sigmoid}_{\lip}$-Lipschitzness of $\sigmoid$ given in \Cref{property:lipschitz}. We then observe that $I_{2} \leq 0.5 I_{1}$ when we choose 
    \begin{equation*}
        \eta \leq \frac{\gamma^{2}}{\norm{\sigmoid}_{\lip}^{2}}\frac{N\lambda_{\min}(\mX_{\TP}\mX_{\TP}^{\T})}{2\lambda_{\max}^{2}(\mX_{\TP}\mX_{\TP}^{\T})}.
    \end{equation*}
    \\\noindent\textbf{Step 2:} Upper bounding the corrupted gradient. We now upper bound the corrupted gradient term. 
    \begin{equation*}
        \begin{aligned}
        III^{2} \overset{\defn}&{=} \frac{4\eta^2}{\bracks*{(1-\epsilon)N}^{2}}\cdot \normsm{\sum_{i \in \FP}(\sigmoid(\vw \cdot \vx_{i})- y_{i})\cdot \sigmoid'(\vw^{(t)} \cdot \vx_{i})\cdot \vx_{i}}_{2}^{2} \\
        &\leq \frac{4\eta^2}{\bracks*{(1-\epsilon)N}^{2}}\cdot \norm{\sigmoid}_{\lip}^{2}\norm{\mX_{\FP}\mX_{\FP}^{\T}}_{2}\sum_{i \in \FP}\bigl(\sigmoid(\vw \cdot \vx_{i}) - \sigmoid(\vw^{*}\cdot\vx_{i})\bigr)^{2} \\
        &\leq \frac{4\eta^2}{\bracks*{(1-\epsilon)N}^{2}}\cdot \norm{\sigmoid}_{\lip}^{2}\norm{\mX_{\FP}\mX_{\FP}^{\T}}_{2}\sum_{i \in \FN}\bigl(\sigmoid(\vw \cdot \vx_{i}) - \sigmoid(\vw^{*}\cdot\vx_{i}) + \xi_{i}\bigr)^{2}\\
        &\leq \frac{8\eta^2}{\bracks*{(1-\epsilon)N}^{2}}\cdot \norm{\sigmoid}_{\lip}^{2}\norm{\mX_{\FP}\mX_{\FP}^{\T}}_{2}\paren*{\norm{\sigmoid}_{\lip}^{2}\norm{\mX_{\FN}\mX_{\FN}^{\T}}_{2}\norm{\vw^{(t)} - \vw^{*}}_{2}^2 + \norm{\vxi_{\FN}}_{2}^{2}}.
        \end{aligned}
    \end{equation*}
    In the above, the first inequality follows from \Cref{lem:hadamard-vector-matrix}, the second inequality follows from the optimality of the Subquantile set, the final inequality follows from the $\norm{\sigma}_{\lip}$-Lipschitzness of $\sigmoid$. Then from \Cref{lem:nonlinear-neuron-noise-bound-indicator}, we have with probability at least $1-\delta$, 
    \begin{equation*}
        II \overset{\defn}{=} \normsm{\sum_{i \in \TP}\xi_{i}\sigmoid'(\vw^{(t)}\cdot\vx_{i})\cdot \vx_{i}}_{2} \lesssim N\norm{\mGamma}\nu\sqrt{\epsilon\log(1/\epsilon)}.
    \end{equation*}
    We now combine Steps 1 and 2 to give the linear convergence result. Noting that $\sqrt{1-2x} \leq 1-x$ when $x \leq 1/2$, we have for $N = \Omega\paren*{\frac{Rd + \log(1/\delta)}{\epsilon}}$, 
    \begin{equation*}
        \begin{aligned}
        \norm{\vw^{(t+1)} - \vw^{*}}_{2} &\leq \norm{\vw^{(t)} - \vw^{*}}_{2}\paren*{1 - \frac{2\gamma^{2}\eta}{(1-\epsilon)N}\cdot \lambda_{\min}(\mX_{\TP}\mX_{\TP}^{\T}) + \frac{2\eta}{(1-\epsilon)N}\cdot \norm*{\mX_{\FP}}_{2}\norm*{\mX_{\FN}}_{2}}\\
        &\quad+ c_{2}\lambda_{\max}(\mSigma)\nu \sqrt{\epsilon\log(1/\epsilon)} + \frac{\sqrt{8}\eta}{(1-\epsilon)N}\cdot\norm{\mX_{\FP}}_{2}\norm{\vxi_{\FN}}_{2},
        \end{aligned}
    \end{equation*}
     where $c_{2}$ is a constant. 
     \noindent\textbf{Step 3: Concentration Bounds.} We will give the relevant probabilistic bounds for the random variables in Steps 1 and 2. From \Cref{lem:sub-gaussian-matrix-eigenvalues}, we have 
     \begin{equation*}
         \norm*{\mX_{\FN}}_{2}\norm*{\mX_{\FP}}_{2} \leq \epsilon\sqrt{\lambda_{\max}(\mSigma)\cdot 10B N\log(1/\epsilon)},
     \end{equation*} 
     with probability at least $1-\delta$ when
     \begin{equation*}
         N \geq \frac{2}{\epsilon}\cdot\paren*{dC_{K}^2 + \frac{\log(2/\delta)}{c_{K}}} \text{ and } \epsilon \leq \frac{1}{60}\cdot\kappa^{-1}(\mSigma)
     \end{equation*}
     From the same Lemma and under the same data conditions we have 
     \begin{equation*}
         \lambda_{\min}(\mX_{\TP}\mX_{\TP}^{\T}) \geq \frac{1}{4}\cdot\lambda_{\min}(\mSigma).
     \end{equation*}
     Then when the corruption rate satisfies
     \begin{equation*}
         \epsilon \leq \frac{\gamma^{2}\lambda_{\min}(\mSigma)}{\sqrt{32B \lambda_{\max}(\mSigma)}},
     \end{equation*}
     we have
     \begin{equation*}
         \norm{\mX_{\FP}}_{2}\norm{\mX_{\FN}}_{2} - \gamma^{2}\lambda_{\min}\paren{\mX_{\TP}\mX_{\TP}^{\T}} \geq \frac{1}{8}\cdot \lambda_{\min}(\mSigma)
     \end{equation*}
     We then have, after $O\paren*{\kappa^2(\mSigma)\log\paren*{\frac{\norm{\vw^{*}}_{2}}{\varepsilon}}}$ iterations with high probability, 
    \begin{equation*}
        \norm{\vw^{(T)} - \vw^{*}}_{2} \leq \varepsilon + O\paren*{C_{\mSigma}C_{\nu}\sqrt{\epsilon\log(1/\epsilon)}} + O\paren*{\gamma^{-2}\kappa(\mSigma)\nu\epsilon\sqrt{B\log(1/\epsilon)}}
    \end{equation*}
    In the final inequality above, we set $\varepsilon = O\paren*{C_{\mSigma}\nu\sqrt{\epsilon\log(1/\epsilon)}}$ for 
    \begin{equation*}
        N = \Omega\paren*{\frac{d + \log(1/\delta)}{\epsilon}}.
    \end{equation*}
    Our proof is complete. 
\end{proof}

\subsection{ReLU Neuron} \label{sec:relu-neuron-error}
In this section, we consider $\relu$ type functions. Our high-level analysis will be similar to the previous sub-sections however the details are considerably different and require stronger conditions we can guarantee by randomness. 
\subsubsection{Proof of \Cref{thm:relu-neuron-error}}\label{sec:one-relu-neuron}
\begin{proof}
    We will now begin our standard analysis.
    \begin{equation*}
        \begin{aligned}
            \norm{\vw^{(t+1)} - \vw^{*}}_{2} &= \norm{\vw^{(t)} - \vw^{*} - \eta \nabla \gR(\vw;\ermS^{(t)})}_{2} \\
            &= \norm{\vw^{(t)} - \vw^{*} - \eta \nabla \gR(\vw^{(t)};\TP)- \eta\nabla\gR(\vw^{(t)};\FP)}_{2} \\
            &\leq \underbrace{\norm{\vw^{(t)} - \vw^{*} - \eta \nabla \gR(\vw^{(t)};\TP)}_{2}}_{I} + \underbrace{\norm{\eta \nabla\gR(\vw^{(t)};\FP)}_{2}}_{II}.
        \end{aligned}
    \end{equation*}
    We will now upper bound $I$ through its square in accordance with our proof sketch, 
    \begin{equation*}
        I^{2} = 
        \norm{\vw^{(t)} - \vw^{*}}_{2}^2 - \underbrace{2\eta \cdot \angle{\vw^{(t)} - \vw^{*}, \nabla \gR(\vw^{(t)};\TP)}}_{I_{1}} + \underbrace{\eta^2 \cdot \norm{\nabla \gR(\vw^{(t)};\TP)}_{2}^{2}}_{I_{2}} .
    \end{equation*}
    We will first lower bound $I_{1}$. We will first adopt the notation from \cite{zhang2019learning}, let $\mSigma_{\TP}(\vw, \hat\vw) = \mX_{\TP}\mX_{\TP}^{\T}\cdot \vone\braces{\mX_{\TP}^{\T}\vw \geq \vzero}\cdot \vone\braces{\mX_{\TP}^{\T}\hat\vw \geq \vzero}$, it then follows
    \begin{equation*}
        \begin{aligned}
            I_{1}
            \overset{\defn}&{=} \frac{4\eta}{(1-\epsilon)N}\cdot \angle{\vw^{(t)} - \vw^{*}, \sum_{i \in \TP}(\sigmoid(\vw^{(t)}\cdot\vx_{i}) - y_i)\cdot \vx_{i} \cdot \vone\braces{\vw^{(t)}\cdot\vx_{i} \geq 0}}\\
            &= \frac{4\eta}{(1-\epsilon)N}\cdot \angle{\vw^{(t)} - \vw^{*}, \mSigma_{\TP}\paren{\vw^{(t)},\vw^{(t)}} \vw^{(t)} - \mSigma_{\TP}\paren{\vw^{(t)},\vw^{*}}\vw^{*}} \\
            &\quad - \frac{4\eta}{(1-\epsilon)N}\cdot \angle{\vw^{(t)} - \vw^{*}, \sum_{i \in \TP}\xi_{i}\vx_{i} \cdot \vone\braces{\vw^{(t)}\cdot \vx_{i} \geq 0}}\\
            &\geq \frac{4\eta}{(1-\epsilon)N}\cdot \angle{\vw^{(t)} - \vw^{*}, \mSigma_{\TP}\paren{\vw^{(t)},\vw^{*}} \paren{\vw^{(t)} - \vw^{*}} + \mSigma_{\TP}\paren{\vw^{(t)},-\vw^{*}}\vw^{(t)}} \\
            &\quad - \frac{4\eta}{(1-\epsilon)N}\cdot \norm{\vw^{(t)} - \vw^{*}}_{2}\normsm{\sum_{i \in \TP}\xi_{i}\vx_{i}\cdot \vone\braces{\vw^{(t)}\cdot \vx_{i} \geq 0}}_{2}\\
            \overset{(i)}&{\geq} \frac{4\eta}{(1-\epsilon)N}\cdot \lambda_{\min}\paren{\mSigma_{\TP}\paren{\vw^{(t)},\vw^{*}}} \norm{\vw^{(t)} - \vw^{*}}_{2}^{2} \\
            &\quad- \frac{4\eta}{(1-\epsilon)N}\cdot \norm{\vw^{(t)} - \vw^{*}}_{2}\normsm{\sum_{i \in \TP}\xi_{i}\vx_{i}\cdot \vone\braces{\vw^{(t)}\cdot \vx_{i} \geq 0}}_{2} \\
            \overset{(ii)}&{\geq} \frac{2\eta}{(1-\epsilon)N}\cdot \lambda_{\min}\paren{\mSigma_{\TP}\paren{\vw^{(t)},\vw^{*}}} \norm{\vw^{(t)} - \vw^{*}}_{2}^{2} \\
            &\quad - \frac{2\eta}{(1-\epsilon)N}\cdot \lambda_{\min}^{-1}\paren{\mSigma_{\TP}(\vw^{(t)},\vw^{*})}\cdot\normsm{\sum_{i \in \TP}\xi_{i}\vx_{i}\cdot \vone\braces{\vw^{(t)}\cdot \vx_{i} \geq 0}}_{2}^{2}.
        \end{aligned}
    \end{equation*}
    In the above, $(ii)$ follows from Young's Inequality and $(i)$ holds from the following relation, 
    \begin{equation*}
        \begin{aligned}
            \angle{\vw^{(t)} - \vw^{*}, &\mSigma_{\TP}(\vw^{(t)},-\vw^{*}) \vw^{(t)}} \nonumber \\
            &= \angle{\vw^{(t)} - \vw^{*}, \sum_{i \in \TP}\vx_{i}\vw^{(t)}\cdot\vx_{i} \cdot \vone\braces{\vw^{(t)}\cdot\vx_{i} \geq 0} \cdot  \vone\braces{\vw^{*}\cdot\vx_{i} \leq 0}}\\
            &= \sum_{i \in \TP} (\vw^{(t)}\cdot\vx_{i} - \vw^{*}\cdot\vx_{i})(\vw^{(t)}\cdot\vx_{i}) \cdot  \vone\braces{\vw^{(t)}\cdot\vx_{i} \geq 0} \cdot \vone\braces{\vw^{*}\cdot\vx_{i} \leq 0} \geq 0.
        \end{aligned}
    \end{equation*}
    In the above, in the final relation we can note that when the indicators are positive, it must follow that both $\vw^{(t)}\cdot\vx_{i}$ is positive and $\vw^{(t)}\cdot\vx_{i} \geq \vw^{*}\cdot\vx_{i}$ as $\vw^{*}\cdot\vx_{i} \leq 0$. We have from Weyl's Inequality,
    \begin{equation*}
        \lambda_{\min}\paren*{\mSigma_{\TP}(\vw^{(t)},\vw^{*})} \geq \lambda_{\min}\paren*{\E\bracks*{\mSigma_{\TP}\paren{\vw^{(t)},\vw^{*}}}} - \norm{\mSigma_{\TP}(\vw^{(t)}, \vw^{*}) - \E\bracks*{\mSigma_{\TP}(\vw^{(t)}, \vw^{*})}}_{2}.
    \end{equation*}
    Let $\Omega = \braces*{\vx \in \sR^{d}: \vx^{\T}\vw^{(t)} \geq 0, \vx^{\T}\vw^{*} \geq 0}$, then
    \begin{equation*}
        \begin{aligned}
            \E_{\vx \sim \gD}\bracks*{\mSigma_{\TP}\paren{\vw^{(t)},\vw^{*}}} &= \sum_{i \in \TP}\E_{\vx \sim \gD}\bracks*{\vx\vx^{\T}\cdot \vone\braces{\vw^{(t)}\cdot\vx_{i} \geq 0}\cdot \vone\braces{\vw^{*}\cdot\vx_{i} \geq 0}} \\
            \overset{(i)}&{\succeq} N(1-2\epsilon)\cdot\paren*{\pi - \Theta^{(t)} - \sin\Theta^{(t)}} \cdot \mI \\
            \overset{(ii)}&{\succeq} N(1-2\epsilon)\cdot \paren*{\pi - 2\arcsin\paren*{\frac{\norm{\vw^{(t)} - \vw^{*}}}{\norm{\vw^{*}}}}}\cdot \mI \\
            &\succeq N(1-2\epsilon)\cdot \pi \paren*{1 - \frac{\norm{\vw^{(t)} - \vw^{*}}}{\norm{\vw^{*}}}}\cdot \mI \\
            &\succsim N(1-2\epsilon)\cdot \mI.
        \end{aligned}
    \end{equation*}
    In the above, $(i)$ follows from \Cref{lem:second-moment-matrix-of-subset-space}, $(ii)$ follows from the guarantee in the randomized initialization.
    We then have from \Cref{lem:halfspace-covariance-estimation}, 
    \begin{equation*}
        \norm{\mSigma_{\TP}(\vw^{(t)}, \vw^{*}) - \E\bracks*{\mSigma_{\TP}(\vw^{(t)}, \vw^{*})}}_{2} \lesssim N\norm{\mGamma}_{2}^{2}\sqrt{\epsilon\log(1/\epsilon)}.
    \end{equation*}
    We then find there exists sufficiently small $\epsilon$ such that the minimimum eigenvalue of $\mSigma_{\TP}(\vw^{(t)},\vw^{*})$ satisfies
    \begin{equation*}
        \lambda_{\min}\paren{\mSigma_{\TP}\paren{\vw^{(t)},\vw^{*}}} \geq \frac{\lambda_{\min}(\mSigma)}{4}.
    \end{equation*}
    We now bound the second-moment matrix approximation. Let $dP(\vx)$ be a Dirac-measure for $\vx \in \TP$. 
    We now upper bound $I_{2}$ by splitting it into two sperate terms, 
    \begin{equation*}
        \begin{aligned}
            I_{2} &= \frac{4\eta^2}{\bracks*{(1-\epsilon)N}^{2}}\cdot \normsm{\sum_{i \in \TP}\paren{\sigmoid(\vw^{(t)}\cdot\vx_{i}) - \sigmoid(\vw^{*}\cdot\vx_{i})  - \xi_{i}}\cdot \vx_{i}\cdot \vone\braces{\vw^{(t)}\cdot\vx_{i} \geq 0}}_{2}^{2} \\
            &\leq \underbrace{\frac{8\eta^2}{\bracks*{(1-\epsilon)N}^{2}}\cdot \normsm{\sum_{i \in \TP}\paren{\sigmoid(\vw^{(t)}\cdot\vx_{i}) - \sigmoid(\vw^{*}\cdot\vx_{i})}\cdot \vx_{i}\cdot \vone\braces{\vw^{(t)}\cdot\vx_{i} \geq 0}}_{2}^2}_{I_{21}} \nonumber \\
            &\quad + \underbrace{\frac{8\eta^2}{\bracks*{(1-\epsilon)N}^{2}}\cdot \normsm{\sum_{i \in \TP}\xi_{i}\vx_{i}\cdot \vone\braces{\vw^{(t)}\cdot\vx_{i}\geq 0}}_{2}^2}_{I_{22}}.
        \end{aligned}
    \end{equation*}
    Recall from \Cref{lem:nonlinear-neuron-noise-bound-indicator}, we have an upper bound on $I_{22}$. We next upper bound $I_{21}$.  
    \begin{equation*}
        \begin{aligned}
            I_{22} &= \frac{8\eta^2}{\bracks*{(1-\epsilon)N}^{2}}\cdot\normsm{\sum_{i \in \TP}\paren{\sigmoid(\vw^{(t)}\cdot\vx_{i}) - \sigmoid(\vw^{*}\cdot\vx_{i})}\cdot \vx_{i}\cdot \vone\braces{\vw^{(t)}\cdot\vx_{i} \geq 0}}_{2}^2 \\
            &\leq \frac{8\eta^2}{\bracks*{(1-\epsilon)N}^2}\cdot \norm*{\mX_{\TP}\mX_{\TP}^{\T}}_{2}\sum_{i \in \TP}\paren{\sigmoid(\vw^{(t)}\cdot\vx_{i}) - \sigmoid(\vw^{*}\cdot\vx_{i})}^{2} \\
            &\leq \frac{8\eta^2}{\bracks*{(1-\epsilon)N}^2}\cdot \norm*{\mX_{\TP}\mX_{\TP}^{\T}}_{2}^{2}\norm{\vw^{(t)} - \vw^{*}}_{2}^2. 
        \end{aligned}
    \end{equation*}
    Then, by choosing $\eta \leq \frac{\lambda_{\min}(\mSigma)}{80\lambda_{\max}^2(\mSigma)}$, we have that $ I_{22} \leq \frac{\lambda_{\min}(\mSigma)}{8}$.  
    \\\noindent\textbf{Step 3: Upper bounding the corrupted gradient.} We now upper bound $II$. 
    \begin{equation*}
        \begin{aligned}
            II &=  \frac{4\eta^2}{\bracks*{(1-\epsilon)N}^{2}}\cdot \normsm{\sum_{i \in \FP}(\sigmoid(\vw^{(t)}\cdot\vx_{i})-y_{i})\cdot \vx_{i} \cdot \vone\braces{\vw^{(t)}\cdot\vx_{i} \geq 0}}_{2}^{2} \\
            &\leq \frac{4\eta^2}{\bracks*{(1-\epsilon)N}^2}\cdot \norm{\mSigma_{\FP}(\vw^{(t)},\vw^{(t)})}_{2}\sum_{i \in \FP}(\sigmoid(\vw^{(t)}\cdot\vx_{i})-\sigmoid\paren* + \xi_{i})^{2} \\
            &\leq \frac{4\eta^2}{\bracks*{(1-\epsilon)N}^2}\cdot \norm{\mX_{\FP}\mX_{\FP}^{\T}}_{2}\sum_{i \in \FN}(\sigmoid(\vw^{(t)}\cdot\vx_{i})-\sigmoid\paren*{\vw^{*}\cdot\vx_{i}} + \xi_{i})^{2}\\
            &\leq \frac{8\eta^2}{\bracks*{(1-\epsilon)N}^2}\cdot \norm{\mX_{\FP}\mX_{\FP}^{\T}}_{2}\paren*{\norm{\mX_{\FN}\mX_{\FN}^{\T}}_{2}\norm{\vw^{(t)} - \vw^{*}}_{2}^{2} + \norm{\vxi_{\FN}}_{2}^{2}}.
        \end{aligned}
    \end{equation*}
    In the above, the first inequality follows from the same argument as \Cref{lem:hadamard-vector-matrix}, the second inequality follows from the optimality of the Subquantile set, and the final inequality follows from noting that $\sigmoid$ is $1$-Lipschitz. 
    We now conclude Steps 1-3 with our linear convergence result. 
    \begin{multline*}
        \norm{\vw^{(t+1)} - \vw^{*}}_{2} \lesssim  \norm{\vw^{(t)} - \vw^{*}}\paren*{1 - \frac{\eta}{32}\cdot \lambda_{\min}(\mSigma) + \frac{4\eta}{(1-\epsilon)N}\cdot \norm*{\mX_{\FP}}_{2}\norm*{\mX_{\FN}}_{2}} 
        \\ + \paren*{\frac{8\eta}{N} + \frac{4\eta}{N}\cdot\sqrt{\lambda_{\min}^{-1}(\mSigma})}\cdot \normsm{\sum_{i \in \TP}\xi_{i}\vx_{i}\cdot\vone\braces{\vw^{(t)}\cdot\vx_{i}\geq 0}}_{2} + \frac{\eta}{(1-\epsilon)N}\norm{\mX_{\FP}}_{2}\norm{\vxi_{\FN}}_{2}.
    \end{multline*}
    \textbf{Step 4: Concentration Inequalities.} From our previous theorems, we have 
    \begin{equation*}
        \norm*{\mX_{\FP}}_{2}\norm*{\mX_{\FN}}_{2} \leq N\epsilon \sqrt{10B \log(1/\epsilon)},
    \end{equation*}
    with probability exceeding $1-\delta$ and $N = \Omega\paren*{\frac{d + \log(2/\delta)}{\epsilon}}$. From \Cref{lem:nonlinear-neuron-noise-bound-indicator}, we have with probability exceeding $1 - \delta$, 
    \begin{equation*}
        \normsm{\sum_{i \in \TP}\xi_{i}\vx_{i}\cdot\vone\braces{\vw^{(t)}\cdot\vx_{i}\geq 0}}_{2} \lesssim \nu\norm{\mGamma}_{2}\sqrt{\epsilon\log(1/\epsilon)}. 
    \end{equation*}
    We furthermore have with failure probability less than $\delta$ from \Cref{lem:chi-squared-sum}, 
    \begin{equation*}
        \norm{\vxi_{\FN}}_{2} \leq \nu\sqrt{30\epsilon\log(1/\epsilon)}.
    \end{equation*}If $\epsilon \leq \frac{1}{64\sqrt{80\log(2)}}$, we obtain after $T = O\paren*{\kappa^2(\mSigma)\log\paren*{\frac{\norm{\vw^{*}}_{2}}{\varepsilon}}}$ gradient descent iterations, 
    \begin{equation*}
        \begin{aligned}
            \norm{\vw^{(t+1)} - \vw^{*}}_{2} &\lesssim  \varepsilon + \nu\norm{\mGamma}_{2}\lambda_{\min}^{-1}(\mSigma)\sqrt{\epsilon\log(1/\epsilon)} + \nu\epsilon\norm{\mGamma}_{2}\sqrt{B\log(1/\epsilon)}\\
            &= O\paren*{\nu\norm{\mGamma}_{2}\lambda_{\min}^{-1}(\mSigma)\sqrt{\epsilon\log(1/\epsilon)}} + O\paren*{\nu\epsilon\norm{\mGamma}_{2}\sqrt{B\log(1/\epsilon)}}.
        \end{aligned}
    \end{equation*}
    when we choose $\varepsilon = O\paren*{\nu\norm{\mGamma}_{2}\lambda_{\min}^{-1}(\mSigma)\sqrt{\epsilon\log(1/\epsilon)}} + O\paren*{\nu\epsilon\norm{\mGamma}_{2}\sqrt{B\log(1/\epsilon)}}$. Our proof is complete. 
\end{proof}
\section{Probability Theory}\label{sec:probability-theory}
In this section, we will present and prove various concentration inequalities and upper bounds for random variables. 
\subsection{$\chi$-Squared Random Variables}
\begin{lemma}[Upper Bound on Sum of Chi-Squared Variables \citep{laurent2000adaptive}]\label{lem:chi-squared-sum}
    Suppose $\xi_{i} \sim \gN(0,\nu^{2})$ for $i \in [n]$, then 
    \begin{equation*}
        \Prob\braces*{\norm{\vxi}_{2}^{2} \geq \nu\paren*{n + 2\sqrt{nx} + 2x}}\leq \e^{-x}.
    \end{equation*}
\end{lemma}
\begin{proposition}[Probabilistic Upper Bound on Sum of Chi-Squared Variables]\label{prop:chi-squared-sum}
    Suppose $\xi_{i} \sim \gN(0,\nu^{2})$ for $i \in [n]$. Let $S \subset [n]$ such that $|S| = \epsilon n$ for $\epsilon \in (0,0.5)$ and let $\gW$ represent all such subsets. Given a failure probability $\delta \in (0,1)$, when $n \geq \log(1/\delta)$, with probability exceeding $1-\delta$,  
    \begin{equation*}
        \max_{\ermS \in \gW}\norm{\vxi_{\ermS}}_{2}^{2} \leq \nu \paren*{30  n \epsilon \log (1/\epsilon)}.
    \end{equation*}
\end{proposition}
\begin{proof}
    Directly from \Cref{lem:chi-squared-sum}, we have with probability exceeding $1- \delta$. 
    \begin{equation*}
        \norm{\vxi}_{2}^{2} \leq \nu\paren*{n + 2\sqrt{n\log(1/\delta)} + 2\log(1/\delta)}
    \end{equation*}
    We now can prove the claimed bound using the union bound, 
    \begin{equation*}
        \Prob\braces*{\max_{\ermS \in \gW}\norm{\vxi}_{2}^{2} \geq \nu\paren*{\epsilon n + 2\sqrt{\epsilon nx} + 2x}} \leq \paren*{\frac{\e}{\epsilon}}^{\epsilon n}\Prob\braces*{\norm{\vxi}_{2}^{2} \geq \nu\paren*{\epsilon n + 2\sqrt{\epsilon nx} + 2x}} \leq \paren*{\frac{\e}{\epsilon}}^{\epsilon n}\e^{-x}.
    \end{equation*}
    In the first inequality we apply a union bound over $\gW$ with \Cref{lem:binom-sum}, and in the second inequality we use \Cref{lem:chi-squared-sum}. We then obtain with probability exceeding $1-\delta$, 
    \begin{equation*}
        \begin{aligned}
        \max_{\ermS \in \gW} \norm{\vxi_{\ermS}}_{2}^{2} &\leq \nu\paren*{\epsilon n + 2\sqrt{n\epsilon \log(1/\delta) + 3n^2\epsilon^2\log(1/\epsilon)}+2\log(1/\delta) + 6n\epsilon\log(1/\epsilon)} \\
        &\leq \nu\paren*{9n\epsilon\log(1/\epsilon) + 2\sqrt{n\epsilon\log(1/\delta)} + 2\sqrt{3}n\epsilon\sqrt{\log(1/\epsilon)} + 2\log(1/\delta)} \\
        &\leq \nu\paren*{15n\epsilon\log(1/\epsilon) + 2\sqrt{n\epsilon \log(1/\delta)} + 2\log(1/\delta)}\\
        &\leq \nu\paren*{30n\epsilon\log(1/\epsilon)}.
        \end{aligned}
    \end{equation*}
    In the above, in the first inequality, we note that $\log\binom{n}{\epsilon n} \leq 3n \epsilon \log(1/\epsilon)$ as $\epsilon < 0.5$, in the second inequality we note that $\sqrt{\log(1/\epsilon)} \leq \paren*{\log(2)}^{-1/2}\log(1/\epsilon) \leq \sqrt{3}\log(1/\epsilon)$ when $\epsilon < 0.5$, the final inequality holds when $n \geq \log(1/\delta)$ by solving for the quadratic equation. The proof is complete. 
\end{proof}
\subsection{Eigenvalue Concentration Inequalities}
In this section, we derive concentration for the minimal and maximal eigenvalues of the sample covariance matrix with $N\epsilon$ samples removed by the adversary. 
\begin{lemma}[Sub-Gaussian Covariance Matrix Estimation \cite{vershynin2010introduction}]\label{lem:vershynin-5-40}
    Let $\mX \in \sR^{d \times n}$ have columns sampled from a sub-Gaussian distribution with sub-Gaussian norm $K$ and second-moment matrix $\mSigma$, then there exists positive constants $c_{k}, C_{K}$, dependent on the sub-Gaussian norm such that with probability at least $1-2\e^{-c_{K}t^{2}}$, 
    \begin{equation*}
        \lambda_{\max}(\mX\mX^{\T}) \leq n\cdot \lambda_{\max}(\mSigma) + \lambda_{\max}(\mSigma) \cdot \paren*{C_{K} \sqrt{dn} + t \sqrt{n}}.
    \end{equation*}
\end{lemma}
\begin{lemma}\label{lem:sub-gaussian-matrix-eigenvalues}
    Let $\mX \in \sR^{d \times n}$ have columns sampled from a sub-Gaussian distribution with sub-Gaussian norm $K$ and second-moment matrix $\mSigma$. Let $S \subset [n]$ such that $|S| = \epsilon n$ for $\epsilon \in (0,0.5)$ and let $\gW$ represent all such subsets. Then with probability at least $1-\delta$, 
    \begin{equation*}
        \begin{aligned}
            &\max_{\ermS \in \gW}\lambda_{\max}(\mX_{\ermS}\mX_{\ermS}^{\T}) \leq \lambda_{\max}(\mSigma)\cdot \paren*{10n\epsilon\log(1/\epsilon)} \\
            &\min_{S \in \gW}\lambda_{\min}(\mX_{[n] \setminus \ermS}\mX_{[n] \setminus \ermS}^{\T}) \geq \frac{n}{4}\cdot \lambda_{\min}(\mSigma)
        \end{aligned}
    \end{equation*}
    when 
    \begin{equation*}
        n \geq \frac{2}{\epsilon}\cdot\paren*{C_{K}^{2} \cdot d + \frac{\log(2/\delta)}{c_{K}}} \;\mathrm{and}\; \epsilon \leq \frac{1}{30} \cdot \kappa^{-1}(\mSigma).
    \end{equation*}
\end{lemma}
\begin{proof}
    We will use a union bound to obtain our claimed error bound. 
    \begin{equation*}
        \begin{aligned}
            \Prob&\braces*{\max_{S \in \gW}\lambda_{\max}(\mX_{\ermS}\mX_{\ermS}^{\T}) \geq n\epsilon \cdot \lambda_{\max}(\mSigma) + \lambda_{\max}(\mSigma)\cdot \paren*{C_{K}\cdot \sqrt{dn\epsilon} + t\sqrt{n\epsilon}}} \\
            &\leq \paren*{\frac{\e}{\epsilon}}^{\epsilon n}\Prob\braces*{\lambda_{\max}(\mX_{\ermS}\mX_{\ermS}^{\T}) \geq n\epsilon \cdot \lambda_{\max}(\mSigma) + \lambda_{\max}(\mSigma)\cdot \paren*{C_{K}\cdot \sqrt{dn\epsilon} + t\sqrt{n\epsilon}}} \\
            &\leq 2 \cdot \paren*{\frac{\e}{\epsilon}}^{\epsilon n}\e^{-c_{K}t^{2}} \leq \delta.
        \end{aligned}
    \end{equation*}
    In the above, the first inequality follows from a union bound over $\gW$ and \Cref{lem:binom-sum}, the second inequality follows from \Cref{lem:vershynin-5-40}. Then from elementary inequalities, we obtain with probability $1 - \delta$, 
    \begin{equation*}\begin{aligned}
        \lambda_{\max}(\mX_{\ermS}\mX_{\ermS}^{\T}) &\leq n\epsilon \cdot \lambda_{\max}(\mSigma) + \lambda_{\max}(\mSigma)\cdot \paren*{C_{K}\cdot \sqrt{dn\epsilon} + \sqrt{\frac{1}{c_{K}}\paren*{n\epsilon\cdot \log(2/\delta) + 3n^2\epsilon^2\log(1/\epsilon)}}}\\
        &\leq n \cdot \lambda_{\max}(\mSigma)\cdot (\epsilon + 3^{3/4}\epsilon\log(1/\epsilon))+ \lambda_{\max}(\mSigma)\cdot \paren*{C_{K}\cdot \sqrt{dn\epsilon} + \sqrt{\frac{1}{c_{K}}n\epsilon \cdot \log(2/\delta)}}\\
        &\leq \lambda_{\max}(\mSigma)\cdot \paren*{6n\epsilon \log(1/\epsilon)} + \lambda_{\max}(\mSigma)\cdot \paren*{{C_{K}\cdot \sqrt{dn\epsilon} + \sqrt{\frac{1}{c_{K}}n\epsilon \cdot \log(2/\delta)}}}\\
        &\leq \lambda_{\max}(\mSigma)\cdot \paren*{10n\epsilon\log(1/\epsilon)}.
    \end{aligned}\end{equation*}
    In the above, the last inequality holds when 
    \begin{equation*}
        n \geq \frac{2}{\epsilon}\cdot\paren*{C_{K}^{2} \cdot d + \frac{\log(2/\delta)}{c_{K}}}.
    \end{equation*}
    and our proof of the upper bound for the maximal eigenvalue is complete. 
    We have from Weyl's Inequality for any $\ermS \in \gW$, 
    \begin{equation*}
        \lambda_{\min}(\mX_{[n]\setminus\ermS}\mX_{[n]\setminus\ermS}^{\T}) = \lambda_{\min}\paren{\mX\mX^{\T} - \mX_{\ermS}\mX_{\ermS}^{\T}} \geq \lambda_{\min}\paren{\mX\mX^{\T}} - \lambda_{\max}\paren{\mX_{\ermS}\mX_{\ermS}^{\T}}
    \end{equation*}
    We then have with probability at least $1-\delta$, 
    \begin{equation*}\begin{aligned}
        \lambda_{\min}(\mX_{[n]\setminus\ermS}\mX_{[n]\setminus\ermS}^{\T}) &\geq n\cdot \lambda_{\min}(\mSigma) - C_{K}\cdot \sqrt{dn} - \sqrt{\frac{1}{c_{K}}\cdot n\cdot \log(2/\delta)} - \lambda_{\max}(\mSigma)\cdot (10n\epsilon \log(1/\epsilon)) \\
        &\geq \frac{3n}{4}\cdot  \lambda_{\min}(\mSigma) - \lambda_{\max}(\mSigma)\cdot (10n\epsilon\log(1/\epsilon)) \geq \frac{n}{4}\cdot \lambda_{\min}(\mSigma).
    \end{aligned}\end{equation*}
    In the above, the first inequality follows when $n \geq \frac{32}{\lambda_{\min}^{2}(\mSigma)}\paren*{C_{K}\cdot d + \frac{1}{c_{K}}\cdot \log(2/\delta)}$, and from some algebra, we find the last inequality holds when $\epsilon \leq \frac{1}{30}\cdot \kappa^{-1}(\mSigma)$ by noting that $\epsilon < 0.5$. The proof is complete. 
\end{proof}
\subsection{Sum of product of nonlinear random variables}
We will first define the Orlicz norm for sub-Gaussian random variables. 
\begin{definition}\label{defn:sub-gaussian-norm}
    The sub-Gaussian norm of a random variable $X$ is denoted as $\norm{X}_{\psi_{2}}$, and is defined as 
    \begin{equation*}
        \norm{X}_{\psi_{2}} = \inf\braces*{t > 0: \E\bracks*{\exp\paren*{X^{2}/t^{2}}} \leq 2}. 
    \end{equation*}
\end{definition}
We first note that Gaussian scalars are sub-Gaussian and $\norm{X}_{\psi_{2}} = O\paren{\nu}$ for $X \sim \gN(0,\nu^{2})$. 
\begin{lemma}\label{lem:sub-gaussian-norm-gaussian-rv}
    Let $X \sim \gN(0, \nu^2)$, then $\norm{X}_{\psi_{2}} = \sqrt{8/3}\nu$. 
\end{lemma}
\begin{proof}
    We have from \Cref{defn:sub-gaussian-norm} that
    \begin{equation*}
        \norm{X}_{\psi_{2}} = \inf\braces*{c \geq 0: \E\bracks*{\exp\paren{X^2/c^2}} \leq 2}.
    \end{equation*}
    We will now solve for the minimizing $c$. From the PDF of a standard Gaussian, 
    \begin{equation*}
        \begin{aligned}
            \E\bracks*{\exp\paren{X^2/c^2}} &= \frac{1}{\nu\sqrt{2\pi}}\int_{-\infty}^{\infty}\exp\paren*{X^2\paren*{\frac{1}{c^2} - \frac{1}{2\nu^{2}}}}dX\\
            &= \frac{1}{\nu\sqrt{2}}\paren*{\frac{1}{c^2} - \frac{1}{2\nu^{2}}}^{-1/2} = 2. 
        \end{aligned}
    \end{equation*}
    From some algebra, we find the final inequality above holds when $c = \sqrt{8/3}\nu$. 
\end{proof}
We now will define the Orlicz norm for sub-exponential random variables. 
\begin{definition}\label{defn:sub-exponential-norm}
    The sub-exponential norm of a random variable $X$ is denoted as $\norm{X}_{\psi_{1}}$, and is defined as 
    \begin{equation*}
        \norm{X}_{\psi_{1}} = \inf\braces*{t > 0: \E\bracks*{\exp\paren*{\abs{X}/t}} \leq 2}. 
    \end{equation*}
\end{definition}
Sub-exponential random variables appear frequently in our analysis from the consequence of the following lemma. 
\begin{lemma}[Lemma 2.7.7 in \citet{vershynin2020high}]\label{lem:sub-Gaussian-product}
    Let $X,Y$ be sub-Gaussian random variables, then $XY$ is sub-exponential, furthermore, 
    \begin{equation*}
        \norm{XY}_{\psi_{1}} \leq \norm{X}_{\psi_{2}}\norm{Y}_{\psi_{2}}
    \end{equation*}
\end{lemma}
To give probabilistic bounds on the concentration of sub-exponential random variables, we often utilize Bernstein's Theorem. 
\begin{lemma}[Proposition 5.16 in \cite{vershynin2010introduction}]\label{lem:verhsynin-5.16}
    Let $X_{1},\ldots, X_{N}$ be independent centered sub-exponential random variables, and $K = \max_{i}\norm{X_{i}}_{\psi_{1}}$. Then for every $\va \in \sR^{n}$ and $t \geq 0$, 
    \begin{equation*}
        \Prob\bracesm{\absm{\sum_{i \in [N]}a_{i}X_{i}} \geq t} \leq 2\exp\bracks*{-c \min\paren*{\frac{t^2}{K^2\norm{\va}_{2}^{2}}, \frac{t}{K\norm{\va}_{\infty}}}}.
    \end{equation*}
\end{lemma}
We are now ready to prove our main results of the section. We use \Cref{lem:nonlinear-noise-term-1,lem:nonlinear-neuron-noise-term-2-lipschitz} in our proof for \Cref{lem:nonlinear-neuron-noise-bound-lipschitz}. We use \Cref{lem:nonlinear-noise-term-1,prop:nonlinear-noise-term-2-indicator} in our proof for \Cref{lem:nonlinear-neuron-noise-bound-indicator}. 

\begin{lemma}\label{lem:nonlinear-noise-term-1}
    Let $\mX = \bracks{\vx_{1},\ldots,\vx_{N}}$ be the data matrix such that for $i \in \bracks{N}$, $\vx_{i}$ are sampled from a sub-Gaussian distribution with second-moment matrix $\mSigma$ with sub-Gaussian proxy $\mGamma$ and $\xi_{i}$ are sampled from $\gN(0,\nu^2)$. Assume $f:\sR^{d} \mapsto \sR$ is bounded over $\sR$. Let $\gW$ represent all subsets of $[N]$ of size empty to $(1-\epsilon)N$ and $\gN_{1}$ be a $\varepsilon$-cover of $\gS^{d-1}$, then for any $\vu \in \gS^{d-1}$ and $\ermS \in \gW$. With probability at least $1-\delta$,
    \begin{equation*}
        \normsm{\sum_{i \in \ermS}\xi_{i}f(\vx_{i}\cdot\vu)\vx_{i}} 
        \lesssim \nu\norm{\mGamma}_{2}\norm{f}_{\infty}N \sqrt{\epsilon\log(1/\epsilon)},
    \end{equation*}
    where $c$ is an absolute constant and the sample size satisfies 
    \begin{equation*}
        N = \Omega\paren*{\frac{Rd + \log(1/\delta)}{\epsilon}}.
    \end{equation*}
\end{lemma}
\begin{proof}
    We will use the following characterization of the spectral norm. 
    \begin{equation*}
        \normsm{\sum_{i \in \TP}f(\vw \cdot \vx_{i})\xi_{i}\vx_{i}}_{2} = \max_{\vv \in \sS^{d-1}}\bigl|\sum_{i \in \TP}f(\vw \cdot \vx_{i})\xi_{i}\vx_{i} \cdot \vv\bigr|
    \end{equation*}
    We will first show that $f(\vw \cdot \vx_{i})\vx_{i}$ is sub-Gaussian. We first note for any $\vv \in \sS^{d-1}$, the random variable, $\vx_{i} \cdot \vv$ is sub-Gaussian by definition. We then have,
    \begin{equation*}
        \paren*{\E_{\vx \sim \gD}\abs{f(\vw \cdot \vx)\vx \cdot \vv}^{p}}^{1/p} \overset{(i)}{\leq} \paren*{\E_{\vx \sim \gD}\abs{f(\vw \cdot \vx)}^{2p}\E_{\vx \sim \gD}\abs{\vx\cdot \vv}^{2p}}^{1/2p} \overset{(ii)}{\leq} \paren*{\norm{f}_{\infty} \norm{\mGamma}_{2}\sqrt{2}}\sqrt{p}.
    \end{equation*}
    In the above, $(i)$ follows from H\"older's Inequality, $(ii)$ follows from noting from letting $q = 2p$ and noting from \Cref{defn:sub-gaussian} that $\norm{\vx_{i} \cdot \vv}_{L_{q}}$ is upper bounded by $\norm{\mGamma}_{2}\sqrt{q}$. We thus have $f(\vw \cdot \vx_{i})\vx_{i} \cdot \vv$ is sub-Gaussian for any $\vw \in \sR^{d}$ and $\norm{f(\vw \cdot \vx_{i})\vx_{i} \cdot \vv}_{\psi_{2}} \lesssim  \norm{\mGamma}_{2}\norm{f}_{\infty}$. We have $\norm{\xi_{i}}_{\psi_{2}} = \sqrt{8/3}\nu$ from \Cref{lem:sub-gaussian-norm-gaussian-rv}, then from \Cref{lem:sub-Gaussian-product}, the random variable $\xi_{i}f(\vw \cdot \vx_{i})\vx_{i} \cdot \vv$ is sub-exponential s.t. $\norm{\xi_{i}f(\vw \cdot \vx_{i})\vx_{i} \cdot \vv}_{\psi_{1}} \lesssim \norm{\mGamma}_{2}\norm{f}_{\infty} \nu $. Let $\widetilde\vw \in \gN_{1}$ such that $\widetilde\vw = \argmin_{\vu \in \gN_{1}}\norm{\vw - \vu}_{2}$, where $\gN_{1}$ is a $\varepsilon$-cover of $\gB(\vzero, R)$. Let $\gN_{2}$ be a $\varepsilon$-net of $\sS^{d-1}$ such that for any $\vv \in \sS^{d-1}$, there exists $\vu \in \gN_{2}$ such that $\norm{\vu - \vv}_{2} \leq \varepsilon$. Let $\vu^{*} = \argmax_{\vu \in \gN_{2}}\abs{(\vxi \circ f(\mX^{\T}\widetilde\vw))^{\T}\mX\vu}$ and $\vv^{*} = \argmax_{\vv \in \sS^{d-1}}\abs{(\vxi \circ f(\mX^{\T}\widetilde\vw))^{\T}\mX\vv}$. We then have from the triangle inequality,  
    \begin{equation*}
        \begin{aligned}
        \abs{(\vxi \circ f(\mX^{\T}\widetilde\vw))^{\T}\mX\vv^{*} - (\vxi \circ f(\mX^{\T}\widetilde\vw))^{\T}\mX\vu^{*}} &\leq \norm{(\vxi \circ f(\mX^{\T}\widetilde\vw))^{\T}\mX}_{2}\norm{\vu^{*} - \vv^{*}}_{2} \\
        &\leq \varepsilon \cdot \norm{(\vxi \circ f(\mX^{\T}\widetilde\vw))^{\T}\mX}_{2}.
        \end{aligned}
    \end{equation*}
    where in the final inequality we use the definition of a $\varepsilon$-net. 
    We then have from the reverse triangle inequality, 
    \begin{equation*}\begin{aligned}
        \abs{(\vxi \circ f(\mX^{\T}\widetilde\vw))^{\T}\mX\vu^{*}} &\geq \abs{(\vxi \circ f(\mX^{\T}\widetilde\vw))^{\T}\mX\vv^{*}} -  \abs{(\vxi \circ f(\mX^{\T}\widetilde\vw))^{\T}\mX\vu^{*} - (\vxi \circ f(\mX^{\T}\widetilde\vw))^{\T}\mX\vv^{*}}\\
        &\geq (1 - \varepsilon)\abs{(\vxi \circ f(\mX^{\T}\widetilde\vw))^{\T}\mX\vv^{*}}.
    \end{aligned}\end{equation*}
    From rearranging, we obtain 
    \begin{equation}
        \abs{(\vxi \circ f(\mX^{\T}\widetilde\vw))^{\T}\mX\vv^{*}} \leq \frac{1}{1-\varepsilon}\cdot \abs{(\vxi \circ f(\mX^{\T}\widetilde\vw))^{\T}\mX\vu^{*}}. \label{eq:vector-epsilon-net-relation}
    \end{equation}
    With this result we are ready to make the probabilistic bounds. Suppose $\gW$ represents all subsets of $[N]$ of size empty to $(1-\epsilon)N$. Suppose $\gN_{2}$ is a $1/2$-net of $\sS^{d-1}$ and $\gN_{1}$ is a $1/2$-net of $\gB(\vw^{*}, R)$, we can then note that $\norm{\vw^{*}} \leq R$. Then,  
    \begin{equation*}\begin{aligned}
        \Prob&\braces*{\max_{\ermS \in \gW}\max_{\vw \in \gN_{1}}\norm{(\vxi_{\ermS}\circ f(\mX_{\ermS}^{\T}\vw))^{\T}\mX_{\ermS}} \geq t}  \nonumber \\
        \overset{\eqref{eq:vector-epsilon-net-relation}}&{\leq} \Prob\braces*{\max_{\ermS \in \gW}\max_{\vv \in \gN_{2}}\max_{\vw \in \gN_{1}}\abs{(\vxi_{\ermS} \circ f(\mX_{\ermS}^{\T}\vw))^{\T}\mX_{\ermS}\vv} \geq 2t}\\
        \overset{(iii)}&{\leq} 2\cdot6^{Rd+d}\paren*{\frac{\e}{\epsilon}}^{N\epsilon}\exp\bracksm{-c \min\parenm{\frac{t^2}{c_{1}^{2}\nu^2\norm{\mGamma}_{2}^{2}\norm{f}_{\infty}^{2}\abs{\ermS}}, \frac{t}{c_{1}\nu\norm{\mGamma}_{2}\norm{f}_{\infty}}}} \leq \delta
    \end{aligned}\end{equation*}
    In the above, $(iii)$ follows from a union bound over $\gW$, $\gN_{1}$, and $\gN_{2}$, and then applying Bernstein's Inequality (see \Cref{lem:verhsynin-5.16}). 
    We also note that $\log\binom{N}{(1-\epsilon)N} = \log\binom{N}{\epsilon N}$, and then applying \Cref{lem:binom-sum} gives the inequality. Then to satisfy the above probabilistic condition, it must hold that
    \begin{equation*}
        t \geq \sqrt{4/3}c^{-1}c_{1}\nu\norm{\mGamma}_{2}\norm{f}_{\infty} \paren*{2dN\log\paren*{6} + 2NRd\log\paren*{6} + 2N\log(2/\delta) + 6N^{2}\epsilon\log(1/\epsilon)}^{1/2}
    \end{equation*}
    Then, when the sample size satisfies
    \begin{equation*}
        N \geq \frac{(Rd+d)\log(6) + \log(2/\delta)}{3\epsilon} = \Omega\paren*{\frac{Rd + \log(1/\delta)}{\epsilon}}.
    \end{equation*}
    We obtain for any $\ermS \in \gW$ and $\vu \in \gN_{1}$, 
    \begin{equation*}
        \normsm{\sum_{i \in \TP}f(\vu \cdot \vx_{i})\xi_{i}\vx_{i}}_{2} \lesssim \nu\norm{\mGamma}_{2}\norm{f}_{\infty}N\sqrt{\epsilon\log(1/\epsilon)}.
    \end{equation*}
    Our proof is complete. 
\end{proof}
\begin{lemma}\label{lem:nonlinear-neuron-noise-term-2-lipschitz}
    Let $\mX = \bracks{\vx_{1},\ldots,\vx_{N}}$ be the data matrix such that for $i \in \bracks{N}$, $\vx_{i}$ are sampled from a sub-Gaussian distribution with second-moment matrix $\mSigma$ with sub-Gaussian proxy $\mGamma$ and $\xi_{i}$ are sampled from $\gN(0,\nu^2)$. Assume $f:\sR^{d} \to \sR$ is bounded over $\sR$ and Lipschitz. Let $\gW$ represent all subsets of $[N]$ of size empty to $(1-\epsilon)N$. Suppose $\vw \in \gB(\vw^{*},R)$ and $\ermS \in \gW$. Let $\gN_{1}$ be a $\varepsilon$-net of $\sS^{d-1}$ and define $\widetilde\vw = \argmin_{\vu \in \gN_{1}}\norm{\vw - \vu}_{2}$. Set a failure probability $\delta \in (0,1)$, then with probability at least $1-\delta$, 
    \begin{equation*}
        \begin{aligned}
            \max_{\ermS \in \gW}\sup_{\vw \in \gB(\vzero, R)}&\normsm{\sum_{i \in \ermS}\xi_{i}f(\vw \cdot \vx_{i})\vx_{i} - \sum_{i \in \ermS}\xi_{i}f(\widetilde\vw \cdot \vx_{i})\vx_{i}}_{2}  \\
            &\leq \sqrt{N}\norm{f}_{\lip}\norm{\mGamma}_{2}\nu\paren*{\log^{2}(N/\delta) + d\log(N/\delta)}
        \end{aligned}
    \end{equation*}
\end{lemma}
\begin{proof}
    We have from the Cauchy-Schwarz inequality, 
    \begin{equation*}\begin{aligned}
        \sup_{\vw \in \gB(\vzero, R)} &\normsm{\sum_{i \in \ermS}\xi_{i}f(\vw \cdot \vx_{i})\vx_{i} - \sum_{i \in \ermS}\xi_{i}f(\widetilde\vw \cdot \vx_{i})\vx_{i}}_{2} \nonumber \\
        &\leq N\sup_{\vw \in \gB(\vzero, R)}\max_{i \in [N]}\norm{\xi_{i}\vx_{i}}_{2}\norm{f(\vw \cdot \vx_{i}) - f(\widetilde\vw \cdot \vx_{i})}_{2} \\
        &\leq N\norm{f}_{\lip}\sup_{\vw \in \gB(\vzero, R)} \max_{i \in [N]}\norm{\xi_{i}\vx_{i}}_{2}\norm{\paren*{\vw - \widetilde\vw} \cdot \vx_{i}}_{2}\\ 
        &\leq N\norm{f}_{\lip}\varepsilon_{1}\max_{i \in [N]}\norm{\xi_{i}\vx_{i}}_{2}\max_{i \in [N]}\norm{\vx_{i}}_{2}.
    \end{aligned}\end{equation*}
    We will bound the two maximal terms seperately. We first consider the $\norm{\vx_{i}}$ term. We have
    \begin{equation}\label{eq:max-subgaussian-vector-norm}
        \Prob\bracesm{\max_{j \in [N]}\norm{\vx_{j}}_{2} \geq t_{1}} \leq N\Prob\bracesm{2\max_{\vv \in \gN_{2}}\abs{\vx_{j} \cdot \vv} \geq t_{1}} \leq 2N6^{d}\exp\paren*{-\frac{t_{1}^{2}}{2\norm{\mGamma}_{2}}} \leq \delta. 
    \end{equation}
    The final inequality above follows when 
    \begin{equation*}
        t_{1} \geq \sqrt{2\norm{\mGamma}_{2}}\paren*{\log(N) + d\log(6) + \log(2/\delta)}^{1/2}.
    \end{equation*}
    We now consider the term $\xi_{i}\vx_{i}$. We note that for any $\vv \in \gN_{2}$ that $\vx_{i} \cdot \vv$ is sub-Gaussian with norm $\norm{\vx_{i} \cdot \vv}_{\psi_{2}} \lesssim \norm{\mGamma}_{2}$. We have from the PDF of a Gaussian random variable,  
    \begin{equation}\label{eq:max-gaussian}
        \Prob\braces*{\max_{i \in \ermS}\abs{\xi_{i}} \geq t_{2}} \leq \frac{\sqrt{2}N}{\sqrt{\pi}}\int_{t}^{\infty}e^{-\frac{x^2}{2\nu^{2}}}dx \leq \frac{\sqrt{2}N}{\sqrt{\pi}}\int_{t}^{\infty}\frac{xe^{-\frac{x^2}{2\nu^2}}}{\sqrt{t}}dx = \frac{\sqrt{2}N}{\sqrt{\pi}}e^{-\frac{t}{2\nu}}\leq \delta.
    \end{equation}
    The final inequality holds when, 
    \begin{equation*}
        t_{2} \geq 2\nu\paren*{\log(N) + \log(1/\delta)} .
    \end{equation*}    
    Combining our estimates, we have 
    \begin{equation*}
        \Prob_{\vx \sim \gD}\braces*{\max_{i \in [n]}\norm{\xi_{i}\vx_{i}}_{2}\max_{i \in [n]}\norm{\vx_{i}}_{2} \geq t_{1}t_{2}} \leq \delta 
    \end{equation*}
    We then choose $\varepsilon = 1/\sqrt{N}$ and we have with probability at least $1 - \delta$, 
    \begin{equation*}
        \begin{aligned}
            \sup_{\vw \in \gB(\vzero, R)} &\normsm{\sum_{i \in \ermS}\xi_{i}f(\vw \cdot \vx_{i})\vx_{i} - \sum_{i \in \ermS}\xi_{i}f(\widetilde\vw \cdot \vx_{i})\vx_{i}}_{2}\\
            &\lesssim \sqrt{N}\norm{f}_{\lip}\norm{\mGamma}_{2}\nu\paren*{\log^{2}(N/\delta) + d^{1/2}\log(N/\delta)}
        \end{aligned}
    \end{equation*}
    Our proof is complete. 
\end{proof}
\begin{lemma}\label{prop:nonlinear-noise-term-2-indicator}
    Let $\mX = \bracks{\vx_{1},\ldots,\vx_{N}}$ be the data matrix such that for $i \in \bracks{N}$, $\vx_{i}$ are sampled from a sub-Gaussian distribution with second-moment matrix $\mSigma$ with sub-Gaussian proxy $\mGamma$ and $\xi_{i}$ are sampled from $\gN(0,\nu^2)$. Assume $f:\sR^{d} \to \sR$ is of the form,
    \begin{equation*}
        f(x) = \begin{cases}1 & x > 0 \\ c & x \leq 0 \end{cases}.
    \end{equation*} Let $\gW$ represent all subsets of $[N]$ of size empty to $(1-\epsilon)N$. Suppose $\vw \in \gB(\vw^{*},R)$ and $\ermS \in \gW$. Let $\gN_{1}$ be a $\varepsilon$-net of $\sS^{d-1}$ and define $\widetilde\vw = \argmin_{\vu \in \gN_{1}}\norm{\vw - \vu}_{2}$. Set a failure probability $\delta \in (0,1)$, then with probability at least $1-\delta$, 
    \begin{equation*}
        \begin{aligned}
            \max_{\ermS \in \gW}\sup_{\vw \in \gB(\vzero, R)}&\normsm{\sum_{i \in \ermS}\xi_{i}f(\vw \cdot \vx_{i})\vx_{i} - \sum_{i \in \ermS}\xi_{i}f(\widetilde\vw \cdot \vx_{i})\vx_{i}}_{2}\\
            &\lesssim \sqrt{N}\norm{f}_{\lip}\norm{\mGamma}_{2}\nu\paren*{\log^{2}(N/\delta) + d\log(N/\delta)}
        \end{aligned}
    \end{equation*}
\end{lemma}
\begin{proof}
    Recall $\widetilde\vw = \argmin_{\vu \in \gN_{1}}\norm{\vw - \vu}_{2}$ and thus for any $\vw$, we have $\norm{\vw - \widetilde\vw}_{2} \leq \varepsilon_{1}$. We will now consider the two seperate function classes. We have, 
    \begin{equation*}\begin{aligned}
        \sup_{\vw \in \gB(\vzero, R)}&\normsm{\sum_{i \in \ermS}\xi_{i}f(\vw \cdot \vx_{i})\vx_{i} - \sum_{i \in \ermS}\xi_{i}f(\widetilde\vw \cdot \vx_{i})\vx_{i}}_{2}  \\
        &\leq \sup_{\vw \in \gB(\vzero, R)}\max_{i \in [N]}\norm{\xi_{i}\vx_{i}}_{2}\norm{f(\vw \cdot \vx_{i}) - f(\widetilde\vw \cdot \vx_{i})}_{2} \\
        &\leq \max_{i \in [N]}\norm{\xi_{i}\vx_{i}}_{2} \sup_{\vw \in \gB(\vw^*, R)}\vone\braces*{\vw \cdot \vx_{i} \geq 0, \widetilde\vw \cdot \vx_{i} \leq 0} \\
        &\quad + \max_{i \in [N]}\norm{\xi_{i}\vx_{i}}_{2}\sup_{\vw \in \gB(\vw^*, R)}\vone\braces*{\vw\cdot\vx_{i} \leq 0, \widetilde\vw \cdot \vx_{i} \geq 0} 
    \end{aligned}\end{equation*}
    This upper bound holds for both when $f$ is of the form, 
    \begin{equation*}
        f(x) = \begin{cases} 1 & x > 0 \\ c &  x \leq 0 \end{cases}
    \end{equation*}
    for any $c \in [0,1)$. We have with probability at least $1 - \delta$, 
    \begin{equation*}
        \max_{i \in [N]}\norm{\xi_{i}\vx_{i}}_{2} \lesssim \nu\sqrt{\norm{\mGamma}_{2}} \paren*{\log^{3/2}(N/\delta) + d^{1/2}\log(N/\delta)}
    \end{equation*}
    We next consider the second term, 
    \begin{equation*}
        \begin{aligned}
            \Prob_{\vx \sim \gD}&\bracesm{\sup_{\vw \in \gB(\vw^*, R)}\vone\braces*{\vw\cdot\vx_{i} \leq 0, \widetilde\vw \cdot \vx_{i} \geq 0} \geq t_{2}} \\
            &\leq t_{2}^{-1}\E_{\vx \sim \gD}\bracks*{\sup_{\vw \in \gB(\vw^{*},R)}\vone\braces*{\vw\cdot\vx_{i} \leq 0, \widetilde\vw \cdot \vx_{i} \geq 0}} \\
            &\leq \frac{\Theta(\vw, \widetilde\vw)}{2\pi t_{2}} \leq \frac{1}{2\pi t_{2}}\arcsin\paren*{\frac{\norm{\vw - \widetilde\vw}}{\norm{\vw}}} 
            \leq \frac{1}{2\pi t_{2}}\frac{\varepsilon_{1}}{\norm{\vw^{*}} - r} \leq t_{2}^{-1}\varepsilon_{1}c_{1},
        \end{aligned}
    \end{equation*}
    where $c_{1}$ is a constant and $r < R$. We obtain similarly that
    \begin{equation*}
        \Prob_{\vx \sim \gD}\bracesm{\sup_{\vw \in \gB(\vw^*, R)}\vone\braces*{\vw\cdot\vx_{i} \geq 0, \widetilde\vw \cdot \vx_{i} \leq 0} \geq t_{2}} \leq t_{2}^{-1}\varepsilon_{1}c_{1}.
    \end{equation*}
    Now we can combine our estimates and obtain with probability exceeding $1 - \delta$, 
    \begin{equation*}
        \begin{aligned}
            \max_{\ermS \in \gW}\sup_{\vw \in \gB(\vzero, R)}&\normsm{\sum_{i \in \ermS}\xi_{i}f(\vw \cdot \vx_{i})\vx_{i} - \sum_{i \in \ermS}\xi_{i}f(\widetilde\vw \cdot \vx_{i})\vx_{i}}_{2}\\
            &\lesssim \varepsilon_{1}\nu\sqrt{\norm{\mGamma}_{2}}\delta^{-1} \paren*{\log^{3/2}(N/\delta) + d^{1/2}\log(N/\delta)},
        \end{aligned}
    \end{equation*}
    Our proof is complete. 
\end{proof}
\begin{proposition}\label{lem:nonlinear-neuron-noise-bound-lipschitz}
    Let $\mX = \bracks{\vx_{1},\ldots,\vx_{N}}$ be the data matrix such that for $i \in \bracks{N}$, $\vx_{i}$ are sampled from a sub-Gaussian distribution with second-moment matrix $\mSigma$ with sub-Gaussian proxy $\mGamma$ and $\xi_{i}$ are sampled from $\gN(0,\nu^2)$. Assume $f:\sR^{d} \to \sR$ is bounded over $\sR$ and Lipschitz. Let $\gW$ represent all subsets of $[N]$ of size empty to $(1-\epsilon)N$. Suppose $\vw \in \gB(\vw^{*},R)$ and $\ermS \in \gW$. Set a failure probability $\delta \in (0,1)$, then with probability at least $1-\delta$, 
    \begin{equation*}
        \norm{(\vxi_{\ermS} \circ f(\mX_{\ermS}^{\T}\vw))^{\T}\mX_{\ermS}}_{2} \lesssim \nu\norm{\mGamma}_{2}\norm{f}_{\infty}N \sqrt{\epsilon\log(1/\epsilon)}
    \end{equation*}
\end{proposition}
\begin{proof} 
    We use the decomposition given in \citet[Lemma A.4]{zhang2019learning}. We obtain, 
    \begin{equation}\label{eq:noise-two-terms}
        \begin{aligned}[b]
            \Prob&\bracesm{\max_{\ermS \in \gW}\sup_{\vw \in \gB(\vzero, R)}\normsm{\sum_{i \in \ermS}\xi_{i}f(\vw \cdot \vx_{i})\vx_{i}} \geq t}
            \leq \Prob\bracesm{\max_{\ermS \in \gW}\max_{\vu \in \gN_{1}}\normsm{\sum_{i \in \ermS}\xi_{i}f(\vx_{i} \cdot \vu)\vx_{i}}_{2} \geq \frac{t}{2}} \\
            &+ \Prob\bracesm{\max_{\ermS \in \gW}\sup_{\vw \in \gB(\vzero, R)}\normsm{\sum_{i \in \ermS}\xi_{i}f(\vw \cdot \vx_{i})\vx_{i} - \sum_{i \in \ermS}\xi_{i}f(\widetilde\vw \cdot \vx_{i})\vx_{i}}_{2} \geq \frac{t} {2}}.
        \end{aligned}
    \end{equation}
    From \Cref{lem:nonlinear-noise-term-1}, we have the first term of \Cref{eq:noise-two-terms} is less than $\delta/2$ when 
    \begin{equation*}
        t \gtrsim \nu\norm{\mGamma}_{2}\norm{f}_{\infty}N\sqrt{\epsilon\log(1/\epsilon)}.
    \end{equation*} 
    From \Cref{lem:nonlinear-neuron-noise-term-2-lipschitz}, we have the second term of \Cref{eq:noise-two-terms} is less than $\delta/2$ when 
    \begin{equation*}
        t \gtrsim \sqrt{N}\norm{f}_{\lip}\norm{\mGamma}_{2}\nu\paren*{\log^{2}(N/\delta) + d\log(N/\delta)}
    \end{equation*}
    Then, when the sample complexity satisfies 
    \begin{equation*}
        N = \Omega\paren*{\frac{\log^4(N/\delta) + d\log^{2}(N/\delta)}{\epsilon}}
    \end{equation*}
    We then obtain with probability exceeding $1 - \delta$, 
    \begin{equation*}
        \norm{(\vxi_{\ermS} \circ f(\mX_{\ermS}^{\T}\vw))^{\T}\mX_{\ermS}}_{2} \lesssim \nu\norm{\mGamma}_{2}\norm{f}_{\infty}N \sqrt{\epsilon\log(1/\epsilon)}
    \end{equation*}
    Our proof is complete. 
\end{proof}

\begin{proposition}\label{lem:nonlinear-neuron-noise-bound-indicator}
    Let $\mX = \bracks{\vx_{1},\ldots,\vx_{N}}$ be the data matrix such that for $i \in \bracks{N}$, $\vx_{i}$ are sampled from a sub-Gaussian distribution with second-moment matrix $\mSigma$ with sub-Gaussian proxy $\mGamma$ and $\xi_{i}$ are sampled from $\gN(0,\nu^2)$. Assume $f:\sR^{d} \to \sR$ is of the form, 
    \begin{equation*}
        f(x) = \begin{cases}1 & x \geq 0 \\ c & x \leq 0 \end{cases}
    \end{equation*}Let $\gW$ represent all subsets of $[N]$ of size empty to $(1-\epsilon)N$. Suppose $\vw \in \gB(\vw^{*},R)$ and $\ermS \in \gW$. Set a failure probability $\delta \in (0,1)$, then with probability at least $1-\delta$, 
    \begin{equation*}
        \norm{(\vxi_{\ermS} \circ f(\mX_{\ermS}^{\T}\vw))^{\T}\mX_{\ermS}}_{2} \lesssim \nu\norm{\mGamma}_{2}\norm{f}_{\infty}N \sqrt{\epsilon\log(1/\epsilon)}
    \end{equation*}
\end{proposition}
\begin{proof}
    We use the decomposition given in \citet[Lemma A.4]{zhang2019learning}. We obtain, 
    \begin{equation}\label{eq:noise-two-terms}
        \begin{aligned}[b]
            \Prob&\bracesm{\max_{\ermS \in \gW}\sup_{\vw \in \gB(\vzero, R)}\normsm{\sum_{i \in \ermS}\xi_{i}f(\vw \cdot \vx_{i})\vx_{i}} \geq t}
            \leq \Prob\bracesm{\max_{\ermS \in \gW}\max_{\vu \in \gN_{1}}\normsm{\sum_{i \in \ermS}\xi_{i}f(\vx_{i} \cdot \vu)\vx_{i}}_{2} \geq \frac{t}{2}} \\
            &+ \Prob\bracesm{\max_{\ermS \in \gW}\sup_{\vw \in \gB(\vzero, R)}\normsm{\sum_{i \in \ermS}\xi_{i}f(\vw \cdot \vx_{i})\vx_{i} - \sum_{i \in \ermS}\xi_{i}f(\widetilde\vw \cdot \vx_{i})\vx_{i}}_{2} \geq \frac{t} {2}}.
        \end{aligned}
    \end{equation}
    From \Cref{lem:nonlinear-noise-term-1}, we have the first term of \Cref{eq:noise-two-terms} is less than $\delta/2$ when 
    \begin{equation*}
        t \gtrsim \nu\norm{\mGamma}_{2}N\sqrt{\epsilon\log(1/\epsilon)}.
    \end{equation*} 
    From \Cref{lem:nonlinear-neuron-noise-term-2-indicator}, we have the second term of \Cref{eq:noise-two-terms} is less than $\delta/2$ when
    \begin{equation*}
        t \gtrsim \nu\sqrt{\norm{\mGamma}_{2}}\delta^{-1}\paren*{\log^{3/2}(N/\delta) + d^{1/2}\log(N/\delta)}
    \end{equation*}
    Then, when the sample complexity satisfies 
    \begin{equation*}
        N = \Omega\paren*{\frac{\log^{3}(N/\delta) + d\log^2(N/\delta)}{\delta\epsilon}}.
    \end{equation*}
    We have with probability exceeding $1 - \delta$, 
    \begin{equation*}
        \norm{(\vxi_{\ermS} \circ f(\mX_{\ermS}^{\T}\vw))^{\T}\mX_{\ermS}}_{2} \lesssim \nu\norm{\mGamma}_{2}N \sqrt{\epsilon\log(1/\epsilon)}
    \end{equation*}
    Our proof is complete. 
\end{proof}
\subsection{Covariance Matrix Estimation in an intersection of Half-Spaces}
\begin{lemma}\label{lem:halfspace-covariance-estimation}
    Fix $\vw^{*} \in \sR^{d-1}$ and suppose $\vw \in \gB(\vw^{*}, R)$ for a constant $R < \norm{\vw^{*}}$. Sample $\vx_{1},\ldots,\vx_{N}$ i.i.d from a sub-Gaussian distribution with second-moment matrix $\mSigma$ and sub-Gaussian norm $\norm{\mGamma}_{2}$. Suppose $\ermS \subset [N]$ s.t. $|\ermS| \leq (1-\epsilon)N$. Then with probability at least $1-\delta$,    
    \begin{equation*}
        \normsm{\mSigma_{\ermS}\paren{\vw^{(t)},\vw^{*}} - \E_{\vx \sim \gD}\bracks{\mSigma_{\ermS}\paren{\vw^{(t)},\vw^{*}}}}_{2} 
        \lesssim N \norm{\mGamma}\sqrt{\epsilon\log(1/\epsilon)}
    \end{equation*}
\end{lemma}
\begin{proof}
    Let $\gN_{1}$ be an $\varepsilon_{1}$-cover of $\gB(\vw^{*},R)$ and $\gN_{2}$ be an $\varepsilon_{2}$-cover of $\sS^{d-1}$. Let $\widetilde\vw = \argmin_{\vv \in \gN_{1}}\norm{\vw - \vv}_{2}$ throughout the relations. We will use the decomposition given in Theorem 1 of \cite{mei2016landscape} to obtain 
    \begin{equation*}\begin{aligned}
       \Prob\biggl\{\max_{\ermS \in \gW}\sup_{\vw \in \gB(\vw^{*};R)}&\norm{\mSigma_{\ermS}(\vw,\vw^{*}) - \E_{\vx \sim \gD}\bracks{\mSigma_{\ermS}(\vw, \vw^{*})}}_{2} \geq t\biggr\} \\
       &\leq \Prob\braces*{\max_{\ermS \in \gW}\sup_{\vw \in \gB(\vw^{*}; R)} \norm{\mSigma_{\ermS}(\vw, \vw^{*}) - \mSigma_{\ermS}(\widetilde\vw, \vw^{*})}_{2} \geq \frac{t}{3}} \\
       &\quad+ \Prob\braces*{\max_{\ermS \in \gW}\max_{\tilde\vw \in \gN_{1}}\norm{\mSigma_{\ermS}(\widetilde\vw, \vw^{*}) - \E_{\vx \sim \gD}\bracks{\mSigma_{\ermS}(\widetilde\vw, \vw^{*})}}_{2} \geq \frac{t}{3}}\\
       &\quad+ \Prob\bracesm{\sup_{\vw \in \gB(\vw^{*}; R)}\norm{\E_{\vx \sim \gD}\bracks{\mSigma_{\ermS}(\widetilde\vw, \vw^{*})} - \E_{\vx \sim \gD}\bracks{\mSigma_{\ermS}(\vw, \vw^{*})}}_{2} \geq \frac{t}{3}}
    \end{aligned}\end{equation*}
     We bound all terms separately. For the first term, 
     \begin{equation*}\begin{aligned}
        \max_{\ermS \in \gW}\sup_{\vw \in \gB(\vw^{*}; R)}\norm{&\mSigma_{\ermS}(\vw, \vw^{*}) - \mSigma_{\ermS}(\widetilde\vw, \vw^{*})}_{2} \nonumber\\
        &\leq \max_{\ermS \in \gW}\sup_{\vw \in \gB(\vw^{*}; R)}\norm{\mSigma_{\ermS}(\vw, -\widetilde\vw)}_{2} + \max_{\ermS \in \gW}\sup_{\vw \in \gB(\vw^{*}; R)}\norm{\mSigma_{\ermS}(-\vw, \widetilde\vw)}_{2} \\
        &\leq 2\max_{i \in [N]}\norm{\vx_{i}\vx_{i}^{\T}}_{2}\arcsin\paren*{\frac{\norm{\widetilde\vw - \vw}_{2}}{\norm{\vw}_{2}}}
        \leq \varepsilon_{1}\cdot \frac{\pi}{\norm{\vw^{*}}- R}\cdot \max_{i \in [N]}\norm{\vx_{i}\vx_{i}^{\T}}
     \end{aligned}\end{equation*}
     Then from \Cref{eq:max-subgaussian-vector-norm}, we have with probability exceeding $1 - \delta$, 
     \begin{equation*}
        \max_{i \in [N]}\norm{\vx_{i}\vx_{i}^{\T}}_{2} \leq 2\norm{\mGamma}_{2}\paren*{\log(N) + d\log(6) + \log(2/\delta)}.
     \end{equation*}
     For the second term, let $\widetilde\vx = \vx\cdot \vone \braces{\widetilde\vw\cdot\vx \geq 0}\cdot \vone \braces{\vw^{*}\cdot\vx \geq 0}$. We then have, 
     \begin{equation*}\begin{aligned}
         \Prob&\braces*{\max_{\ermS \in \gW} \max_{\tilde\vw \in \gN_{1}}\norm{\mSigma_{\ermS}(\widetilde\vw, \vw^{*}) - \E_{\vx \sim \gD}\bracks*{\mSigma_{\ermS}(\widetilde\vw, \vw^{*})}}_{2} \geq \frac{t}{3}} \\
         &\leq \Prob\braces*{\max_{\ermS \in \gW}\max_{\tilde\vw \in \gN_{1}}\max_{\vv \in \gN_{2}}\abs{\norm{\widetilde\mX_{\ermS}^{\T}\vv}_{2}^{2} - \E_{\vx \sim \gD}\norm{\widetilde\mX_{\ermS}^{\T}\vv}_{2}^{2}} \geq \frac{2t}{3}} \\
         \overset{(i)}&{\leq} 2\paren*{\frac{\e}{\epsilon}}^{N\epsilon}\paren*{3/\varepsilon_{1}}^{Rd}12^{d}\exp\bracks*{-c\min\paren*{\frac{t^2 }{144\cdot 256\norm{\mGamma}_{2}^{2}\abs{\ermS}}, \frac{t}{4\cdot 48\norm{\mGamma}_{2}}}} \leq \frac{\delta}{2}
     \end{aligned}\end{equation*}
     In $(i)$ we note from Lemma 1.12 in \cite{rigollet2023high}, that the random variable $\abs{\widetilde\vx \cdot \vv}^{2} - \E_{\vx \sim \gD}\abs{\widetilde\vx \cdot \vv}^{2}$ is sub-exponential and $\norm{\abs{\widetilde\vx \cdot \vv}^{2} - \E_{\vx \sim \gD}\abs{\widetilde\vx \cdot \vv}^{2}}_{\psi_{1}} \leq 16\norm{\mGamma}_{2}$, we can then apply Bernstein's Inequality (see \Cref{lem:verhsynin-5.16}). The probabilistic condition above is then satisfied when, 
     \begin{equation*}
         t \gtrsim \paren*{N\norm{\mGamma}_{2}^{2}\paren*{Rd\log\paren*{3/\varepsilon_{1}} + d\log\paren*{3/\varepsilon_{2}} + 3N\epsilon\log(1/\epsilon) + \log(4/\delta)}}^{1/2}
     \end{equation*}
     We now consider the third term. 
     \begin{equation*}\begin{aligned}
         \Prob&\bracesm{\sup_{\vw \in \gB(\vw^{*}; R)}\norm{\E_{\mX \sim \gD}\bracks{\mSigma_{\ermS}(\widetilde\vw, \vw^{*})} - \E_{\mX \sim \gD}\bracks{\mSigma_{\ermS}(\vw, \vw^{*})}}_{2} \geq \frac{t}{3}} \\
         &\leq \Prob\bracesm{N\sup_{\vw \in \gB(\vw^{*}; R)}\norm{\E_{\vx \sim \gD}\bracks{\vx\vx^{\T}\cdot \paren*{\vone \braces{\widetilde\vw \cdot \vw \geq 0} - \vone \braces{\vw \cdot \vx \geq 0}}\cdot \vone \braces{\vw^{*} \cdot \vx \geq 0}}}_{2} \geq \frac{t}{3}} \\
         \overset{(ii)}&{\leq} \Prob\bracesm{N\sup_{\vw \in \gB(\vw^{*}; R)} \E_{\vx \sim \gD}\bracks*{\norm{\vx\vx^{\T}}_{2}\abs{\vone \braces{\widetilde\vw \cdot \vx \geq 0} - \vone \braces{\vw \cdot \vx \geq 0}}} \geq \frac{t}{3}} \\
         \overset{(iii)}&{\leq} \Prob\bracesm{N\sup_{\vw \in \gB(\vw^{*}; R)}\paren*{\E_{\vx \sim \gD}\norm{\vx\vx^{\T}}_{2}^{2}\E_{\vx \sim \gD}\abs{\vone \braces{\widetilde\vw \cdot \vx \geq 0} - \vone \braces{\vw \cdot \vx \geq 0}}}^{1/2} \geq \frac{2t}{3}} = 0
     \end{aligned}\end{equation*}
     In the above, $(ii)$ follows from first applying Cauchy-Schwarz inequality and then applying Jensen's inequality, and $(iii)$ follows from H\"older's Inequality. Then from $L_{4} \to L_{2}$ hypercontractivity of $\gD$, we have that $\E_{\vx \sim \gD}\norm{\vx}^{4} \leq L\E_{\vx \sim \gD}\norm{\vx}^2 = L\Tr(\mSigma)$. We now consider the second term, 
     \begin{equation*}
         \E_{\vx \sim \gD}\abs{\vone \braces{\widetilde\vw\cdot\vx \geq 0} - \vone \braces{\vw\cdot\vx\geq 0}} \leq \frac{\Theta\paren{\vw, \widetilde\vw}}{\pi}\leq \frac{2}{\pi}\arcsin\paren*{\frac{\norm{\vw - \widetilde\vw}}{\norm{\vw^{*}}- r}} \leq \varepsilon_{1}c_{1}.
     \end{equation*}
     Then we obtain zero probability as indicated in the statement when 
     \begin{equation*}
         t \gtrsim N\sqrt{L\Tr(\mSigma)\varepsilon_1}
     \end{equation*}
     Combining our results, we choose $\varepsilon_{1} = 1/N$ and $\varepsilon_{2} = 1/2$ for sufficiently large $N = \Omega\paren*{\frac{Rd+ \log(1/\delta)}{\epsilon}}$, 
     \begin{equation*}\begin{aligned}
        \max_{\ermS \in \gW}\sup_{\vw \in \gB(\vw^{*}, R)}&\norm{\mSigma_{\ermS}(\vw, \vw^{*}) - \E_{\vx \sim \gD}\bracks*{\mSigma_{\ermS}(\vw, \vw^{*})}}_{2} \lesssim N\norm{\mGamma}_{2}\sqrt{\epsilon\log(1/\epsilon)}
     \end{aligned}\end{equation*}
     Our proof is complete. 
\end{proof}
\begin{lemma}\label{lem:second-moment-matrix-of-subset-space}
    Suppose $\E_{\vx \sim \gD}\bracks*{\vx} = \vzero$ and $\E_{\vx \sim \gD}\bracks*{\vx\vx^{\T}} = \mI$ for $\vx \sim \gD$ where $\gD$ is a rotationally invariant distribution. Fix $\vw_{1}, \vw_{2} \in \sR^{d}$ and define $\Theta = \arccos\paren*{\frac{\vw_{1} \cdot \vw_{2}}{\norm{\vw_{1}}\norm{\vw_{2}}}} \leq \frac{\pi}{2}$. Then,
    \begin{equation*}
        \E_{\vx \sim \gD}\bracks*{\vx\vx^{\T}\cdot \vone\braces*{\vw_{1} \cdot \vx \geq \vzero} \cdot \vone\braces*{\vw_{2} \cdot \vx \geq \vzero}} \succeq \paren*{\frac{\pi - \Theta - \sin\Theta}{2}}\cdot \mI
    \end{equation*}
\end{lemma}
\begin{proof}
    Since we have $\gD$ is an isotropic distribution, we then have the distribution of $\mU\vx$ is isotropic for any unitary $\mU$. Then consider the unitary matrix, 
    \begin{equation*}
        \mU 
        = \begin{bmatrix}
            \frac{\vw_{1}}{\norm{\vw_{1}}} &  \frac{\vw_{2} - \Proj_{\vw_{2}}\vw_{1}}{\norm{\vw_{2} - \Proj_{\vw_{2}}\vw_{1}}} &  \vu_{3} &  \ldots &  \vu_{d}
        \end{bmatrix}
    \end{equation*} 
    where $\vu_{3}, \ldots, \vu_{d}$ represents some orthonormal basis of the complementary subspace to the subspace spanned by $\vw_{1}$ and $\vw_{2}$. 
    Consider the plane spanned by $\vw_{1}$ and $\vw_{2}$. Then, w.l.o.g let $\vw_{1} = (1,0)$ and rotate $\vw_{2}$ such that it is in the first quadrant with angle $\Theta$ from $\vw_{1}$ from noting that $\Theta \leq \frac{\pi}{2}$. 
    \begin{equation*}\begin{aligned}
        \E_{\vx \sim \gD}&\bracks*{\vx\vx^{\T}\cdot \vone\braces*{\vw_{1} \cdot \vx \geq 0} \cdot \vone\braces*{\vw_{2} \cdot \vx \geq 0}} \\
        &= \E_{\mX \sim \gD}\bracks*{\mU^{\T}\mU\vx\vx^{\T}\mU^{\T}\mU\cdot \vone\braces*{\vw_{1} \cdot \vx \geq 0} \cdot \vone\braces*{\vw_{2} \cdot \vx \geq 0}}\\
        &= \mU^{\T}\E_{\mX \sim \gD}\bracks*{\mU\vx\vx^{\T}\mU^{\T}\cdot \vone\braces*{\vw_{1} \cdot \vx \geq 0} \cdot \vone\braces*{\vw_{2} \cdot \vx \geq 0}}\mU
    \end{aligned}\end{equation*}
    In the above, the first equality follows from noting that $\mU$ is unitary, and the second equality follows from the linearity of expectation. 
    Let $\Omega_{\xi} = \braces*{\vx \in \sR^{d} : \vx^{\T}\mU^{\T}\vw_{\xi} \geq 0}$ for $\xi \in \braces{1,2}$. Then there exists constants $\alpha$ and $\beta$ such that  
    \begin{equation*}\begin{aligned}
        \E_{\vx \sim \gD}&\bracks*{\mU\vx\vx^{\T}\mU^{\T}\cdot \vone\braces*{\vw_{1}\cdot\vx \geq \vzero} \cdot \vone\braces*{\vw_{2}\cdot\vx \geq 0}} \\
        &= \E_{\vy \sim \gD}\bracks*{\vy\vy^{\T}\cdot \vone \braces*{y_{1} \geq \vzero} \cdot \vone\braces*{\alpha y_{1} + \beta y_{2} \geq 0}} \\
        &= 
        \begin{bmatrix}
            {\displaystyle \E_{\vy\sim\gD}\bracks*{y_{1}^{2} \cdot \vone \braces*{y_{1} \geq 0} \cdot \vone\braces*{\alpha y_{1} + \beta y_{2} \geq 0}}} & \cdots & {\displaystyle \E_{\vy\sim\gD}\bracks*{y_{1}y_{d} \cdot \vone \braces*{y_{1} \geq 0} \cdot \vone\braces*{\alpha y_{1} + \beta y_{2} \geq \vzero}}}\\
            \vdots & \ddots & \vdots \\
            {\displaystyle \E_{\vy\sim\gD}\bracks*{y_{d}y_{1} \cdot \vone \braces*{y_{1} \geq \vzero} \cdot \vone\braces*{\alpha y_{1} + \beta y_{2} \geq \vzero}}} & \cdots & {\displaystyle \E_{\vy\sim\gD}\bracks*{y_{d}^{2} \cdot \vone \braces*{y_{1} \geq 0} \cdot \vone\braces*{\alpha y_{1} + \beta y_{2} \geq \vzero}}}
        \end{bmatrix}.
    \end{aligned}\end{equation*}
    In the above, the first relation follows from rotational invariance, which gives us that $\mU\mX \overset{d}{=} \mX$ for any unitary $\mU$. Our main tool is a change of coordinates from the euclidean space to $n$-spherical coordinates as all the angles will be independent. We first consider the non-diagonal elements. Suppose $(i,j) \in \bracks{d}\setminus\bracks{2} \times \bracks{d}\setminus\bracks{2}$ and w.l.o.g $i < j$, from the rotational invariance, we have
    \begin{equation*}
        \E_{\mY\sim\gD}\bracks*{\mY_{j}\mY_{i} \cdot \vone \braces*{\mX \in \Omega}} = \E_{\mY\sim\gD}\bracksm{r^{2}\prod_{k_{1} \in [i-1]}\sin(\phi_{k_{1}}) \cos(\phi_{i}) \prod_{k_2 \in [j-1]}\sin\phi_{k_2}\cos\phi_{j}} = 0
    \end{equation*}
    In the above, the final inequality follows from noting that $\int_{0}^{\pi}\cos\theta d\theta = 0$ and if $j = n$ we have $\int_{0}^{2\pi}\sin\theta d\theta = 0$. 
    Then for the diagonal elements, we have $i = j$, and obtain
    \begin{equation*}\begin{aligned}
        \E_{\vx \sim \gD}&\bracks*{\paren{\vx \cdot \vu_{j}}^{2} \cdot \vone \braces*{\vx \in \Omega}} = \E_{\vx \sim \gD}\bracksm{r^2 \prod_{k \in [j-1]}\sin^{2}(\phi_{k}) \cos^{2}(\phi_{i})}\\
        &= \frac{1}{\SA_{j}}\underbrace{\int_{0}^{\infty} r^{2} d\mu(r)}_{I} \underbrace{\int_{-\pi/2 +\Theta}^{\pi/2}\sin^{2}(\phi_{1})d\phi_{1}}_{II}\underbrace{\int_{0}^{2\pi}\cos^{2}(\phi_{j-1})d\phi_{j-1}\prod_{k \in [j-1]\setminus\braces{1}}\int_{0}^{\pi}\sin^{2}(\phi_{k}) d\phi_{k} }_{III}
    \end{aligned}\end{equation*}
    The term of interest here is $II$. Recall $\Theta$ is the angle between $\vw_{1}$ and $\vw_{2}$, we then have 
    \begin{equation*}
        II = \int_{-\pi/2+ \Theta}^{\pi/2}\sin^{2}(\phi_{1})d\phi_{1} = \frac{2\pi - 2\Theta - \sin\paren{2\Theta}}{4}\geq \frac{\pi - 2\Theta}{2}
    \end{equation*}
    Before proceeding, let us note the following, 
    \begin{equation*}
        \E_{\mX \sim \gD}\bracks*{\paren{\mX^{\T}\mU_{j}}^{2}} = \frac{1}{\SA_{j}} \cdot I \cdot III \cdot \int_{-\pi/2}^{\pi/2}\sin^2(\phi_{1})d\phi_{1} = 1
    \end{equation*}
    In the above, the final equality follows from noting $\gD$ is isotropic and thus diagonal covariance elements are unitary. We will now consider the normalizing term, note that $\E_{\mX \sim \gD}\bracks*{\paren{\mX^{\T}\mU_{j}}^{2}} = 1$ as $\gD$ is isotropic. Then, noting that only $II$ is modified when considering $\vone\braces*{\mX \in \Omega}$. We then obtain $S_{j} = I \cdot III \cdot \int_{0}^{\pi}\sin^{2}(\phi_{1})d\phi_{1}$. Then, from rearranging, we have
    \begin{equation*}
        \E_{\vx \sim \gD}\bracks*{\paren{\vx \cdot \vu_{j}}^{2}\cdot \vone \braces*{\mX \in \Omega}} \geq \frac{\pi - 2\Theta}{\pi}
    \end{equation*}
    We will now consider the principal $2 \times 2$ matrix. Then, consider the vector in $\sR^{2}$ in polar coordinates as $(r \cos \Theta, r\sin\Theta)$ where $\Theta$ is uniformly distributed in $[0,2\pi)$ and $r$ is independent. Then, to calculate the expected outer product, we integrate over the two dimensional space of the intersection of $\vw_{1}$ and $\vw_{2}$, 
    \begin{equation*}\begin{aligned}
        &\bracks*{\E_{\vx \sim \gD}\bracks*{\mU\vx\vx^{\T}\mU^{\T}\cdot \vone\braces*{\vx^{\T}\mU^{\T}\vw_{1} \geq \vzero} \cdot \vone\braces*{\vx^{\T}\mU^{\T}\vw_{2} \geq \vzero}}}_{2, 2} \\
        &= \bracks*{\E_{\vy \sim \gD}\bracks*{\vy\vy^{\T}\cdot \vone\braces*{y_{1} \geq \vzero} \cdot \vone\braces*{\alpha y_{1} + \beta y_{2} \geq \vzero}}}_{2, 2} \\
        &= \int_{0}^{\infty} \int_{-\pi/2 + \Theta}^{\pi/2}\begin{pmatrix} \cos \Theta \\ \sin\Theta \end{pmatrix} \paren*{\cos\Theta, \sin\Theta} r d\Theta dr \\
        &= \int_{0}^{\infty} \frac{r}{2} \begin{pmatrix}\pi - \Theta + \sin\Theta\cos\Theta & \sin^2\Theta \\ \sin^2\Theta & \pi - \Theta - \cos\Theta\sin\Theta \end{pmatrix} dr\\
        &= (1/4)\cdot \E_{\vx \sim \gD}\bracks*{r^2}\begin{pmatrix}\pi - \Theta + \sin\Theta\cos\Theta & \sin^2\Theta \\ \sin^2\Theta & \pi - \Theta - \cos\Theta\sin\Theta \end{pmatrix} \\
        &\succeq \paren*{\frac{\pi - \Theta - \sin\Theta}{2}}\cdot \mI
    \end{aligned}\end{equation*}
    Our proof is complete by noting that the planes perpendicular to that of the plane integrated over remain unchanged by the indicator functions and thus have unitary expectation. 
\end{proof}
\subsection{Scaled Gaussian Matrix}
\begin{lemma}\label{lem:max-scaled-gaussian-matrix-norm}
    Fix $\mS \in \sR^{k \times n}, \mT \in \sR^{m \times \ell}$, then sample a matrix $\mG \in \sR^{n \times m}$ with entries sampled i.i.d from $\gN(0, \nu^2)$, then with probability exceeding $1-\delta$, 
    \begin{equation*}
        \norm{\mS\mG\mT}_{\F} \leq \norm{\mS}_{\F}\norm{\mT}_{\F} \cdot \nu\sqrt{2\log\paren*{2nm/\delta}}.
    \end{equation*}
\end{lemma}
\begin{proof}
    We first expand the square of the Frobenius norm to obtain 
    \begin{equation*}
        \norm{\mS\mG\mT}_{\F}^{2} = \sum_{i \in [k]}\sum_{j \in [\ell]} \sum_{k_{1},k_{2} \in [n] \times [m]}\paren*{\mS_{i,k_{1}}\mG_{k_{1},k_{2}}\mT_{k_{2},j}}^{2} \leq  \norm{\mS}_{\F}^{2}\norm{\mT}_{\F}^{2} \max_{i, j \in \bracks{n} \times \bracks{m}}(\mG_{i,j})^{2}.
    \end{equation*}
    It then suffices to bound the maximum value of a Gaussian squared over $N^2$ samples. Then, we have from a union bound and the definition of a Gaussian random variable, 
    \begin{multline*}
        \Prob_{\mG_{i,j} \sim \gN(0,1)}\braces*{\max_{(i,j) \in [n] \times [m]}\mG_{i,j}^2 \geq t} = \Prob_{\mG_{i,j} \sim \gN(0,1)}\braces*{\max_{(i,j) \in [n] \times [m]} \abs{\mG_{i,j}} \geq \sqrt{t}} \\
        \leq \frac{\sqrt{2}nm}{\pi}\int_{\sqrt{t}}^{\infty}\e^{-\frac{x^2}{2\nu^{2}}}dx \leq \frac{\sqrt{2}nm}{\pi}\int_{\sqrt{t}}^{\infty}\frac{x\e^{-\frac{x^2}{2\nu^{2}}}}{\sqrt{t}}dx = \frac{\sqrt{2}nm}{\sqrt{\pi}}\e^{-\frac{t}{2\nu^{2}}}\leq \delta.
    \end{multline*}
    In the above, $(i)$ follows from a union bound. We thus obtain from elementary inequalities, with failure probability at most $\delta$, 
    \begin{equation*}
        \max_{(i,j) \in [n]\times[m]}\mG_{i,j}^{2} \leq \nu^{2} \cdot \paren*{2\log\paren*{2nm/\delta}}.
    \end{equation*}
    Our proof is complete. 
\end{proof}

\section{Mathematical Tools}
In this section, we state additional lemmas referenced throughout the text for completeness. 
\begin{lemma}\label{lem:hadamard-vector-matrix}
    Let $\va, \vb \in \sR^{n}$ and $\mX \in \sR^{p \times n}$, then 
    \begin{equation*}
        \normsm{\sum_{i \in [n]}a_{i}b_{i}\vx_i}^{2} \leq \norm{\va}_{\infty}^{2}\norm{\vb}_{2}^{2}\norm{\mX\mX^{\T}}_{2}.
    \end{equation*}
\end{lemma}
\begin{proof}
    The proof is a simple calculation. Expanding out the LHS, we have
    \begin{equation*}
        \begin{aligned}
            \normsm{\sum_{i \in [n]} a_{i}b_{i} \vx_i}_{2}^2 &= \sum_{i \in [n]}\sum_{j \in [n]}a_{i}a_{j}b_{i}b_{j}\vx_{i}^{\T}\vx_{j} = (\va \circ \vb)^{\T}\mX^{\T}\mX(\va \circ \vb) \\
            &\leq \norm{\va \circ \vb}_{2}^{2}\normsm{\mX^{\T}\mX}_{2} \leq \norm{\va}_{\infty}^2\norm{\vb}_{2}^{2}\normsm{\mX^{\T}\mX}_{2},
        \end{aligned}
    \end{equation*}
    where the final inequality comes from noting 
    \begin{equation*}
        \norm{\va\circ\vb}^{2} = \sum_{i \in [n]}a_{i}^2b_{i}^2 \leq \max_{i \in [n]}a_{i}^2 \cdot \sum_{i \in [n]} b_i^{2} = \norm{\va}_{\infty}^{2}\norm{\vb}_{2}^{2}.
    \end{equation*}
    Our proof is complete.
\end{proof}
\begin{lemma}[Lemma 3.11 in \cite{bubeck2015convex}]\label{lem:bubeck-3.11}
    Let $f$ be $\beta$-smooth and $\alpha$-strongly convex over $\sR^{n}$, then for all $\vx, \vy \in \sR^{n}$,  
    \begin{equation*}
        \left\langle \nabla f(\vx) - \nabla f(\vy), \vx - \vy\right\rangle \geq \frac{\alpha\beta}{\alpha + \beta}\norm{\vx - \vy}^2 + \frac{1}{\alpha + \beta}\norm{\nabla f(\vx) - \nabla f(\vy)}^2.
    \end{equation*}
\end{lemma}
 \begin{lemma}[Sum of Binomial Coefficients \citep{cormen2022introduction}]\label{lem:binom-sum}
    Let $k, n \in \mathbb{N}$ such that $k \leq n$, then
    \begin{equation*}
        \sum_{i=0}^k \binom{n}{i} \leq \left(\frac{en}{k}\right)^{k}.
    \end{equation*}
\end{lemma}
\begin{lemma}[Corollary 4.2.13 in \cite{vershynin2020high}]\label{lem:unit-ball-covering-number}
    The covering number of the $\ell_2$-norm ball $\gB(\vzero;1)$ for $\varepsilon < 0$, satisfies, 
    \begin{equation*}
        \gN(\gB_{\ell_{2}}^{d}(\vzero, 1), \varepsilon) \leq \paren*{\frac{3}{\varepsilon}}^{d}.
    \end{equation*}
\end{lemma}

\clearpage
\end{document}